\documentclass[runningheads]{llncs}
\usepackage{graphicx}

\usepackage{tikz}
\usepackage{comment}
\usepackage{amsmath,amssymb} 
\usepackage{color}
\usepackage{booktabs}
\usepackage[accsupp]{axessibility}  
\usepackage{multirow}
\usepackage{float}

\usepackage{threeparttable}
\usepackage{adjustbox}
\usepackage{listings}
\usepackage{mwe} %
\usepackage{makecell}
\usepackage{color, colortbl}
\usepackage{footmisc}  %
\usepackage{graphicx}
\usepackage{graphbox}

\makeatletter
\newcommand\figcaption{\def\@captype{figure}\caption}
\newcommand\tabcaption{\def\@captype{table}\caption}
\makeatother

\def\etal{\emph{et al.}}
\def\imagenetlt{\textbf{ImageNet-LT}}
\def\imagenetlts{\textbf{ImageNet-LTS}}

\def\awa{\textbf{AWA2}}
\def\awalts{\textbf{AWA2-LTS}}
\def\pacs{\textbf{PACS}}
\def\lt{\textbf{LT}}
\def\ds{\textbf{DS}}
\def\ltds{\textbf{LT-DS}}
\def\pacsodg{\textbf{PACS-ODG}}
\def\best#1{{\bf {#1}}}
\def\sec#1{{\underline{#1}}}
\usepackage{algorithm}
\usepackage{algpseudocode}
\newcommand{\cmark}{\ding{51}}%
\usepackage{pifont}
\newcommand{\accu}{\textbf{\textit{Acc-U}}}
\newcommand{\hu}{\textbf{\textit{H-U}}}
\newcommand{\acc}{\textbf{\textit{Acc}}}

\newcommand{\bx}{\boldsymbol{x}}
\definecolor{Line}{rgb}{.5,.5,1}

\usepackage[colorlinks=true,citecolor=green,urlcolor=red]{hyperref}
\usepackage{subfigure}

\begin{document}

\pagestyle{headings}
\mainmatter
\def\ECCVSubNumber{180}  

\title{Tackling Long-Tailed Category Distribution \\ Under Domain Shifts} 

\titlerunning{Tackling Long-Tailed under Domain Shifts.}
%
\author{Xiao~Gu\inst{1} \and
Yao~Guo\inst{2} \and
Zeju~Li\inst{1} \and
Jianing~Qiu\inst{1} \and
Qi~Dou\inst{3} \and
Yuxuan~Liu\inst{2} \and
Benny~Lo\inst{1}$^*$ \and
Guang-Zhong~Yang\inst{2}$^*$}
\authorrunning{X. Gu et al.}
%
\institute{Imperial College London \\
\email{\{xiao.gu17,zeju.li18,jianing.qiu17,benny.lo\}@imperial.ac.uk} \and \nolinebreak
Shanghai Jiao Tong University \\
\email{\{yao.guo, 20000905lyx, gzyang\}@sjtu.edu.cn} \and \nolinebreak
{The Chinese University of Hong Kong} \\
\email{qidou@cuhk.edu.hk}}
\maketitle
\begin{abstract} 
Machine learning models fail to perform well on real-world applications when 1) the category distribution $P(Y)$ of the training dataset suffers from long-tailed distribution and 2) the test data is drawn from different conditional distributions $P(X|Y)$. Existing approaches cannot handle the scenario where both issues exist, which however is common for real-world applications. In this study, we took a step forward and looked into the problem of long-tailed classification under domain shifts. We designed three novel core functional blocks including Distribution Calibrated Classification Loss, Visual-Semantic Mapping and Semantic-Similarity Guided Augmentation. Furthermore, we adopted a meta-learning framework which integrates these three blocks to improve domain generalization on unseen target domains. Two new datasets were proposed for this problem, named AWA2-LTS and ImageNet-LTS. We evaluated our method on the two datasets and extensive experimental results demonstrate that our proposed method can achieve superior performance over state-of-the-art long-tailed/domain generalization approaches and the combinations. Source codes and datasets can be found at our project page \url{https://xiaogu.site/LTDS}.

\end{abstract}

\section{Introduction}

Deep learning has made unprecedented achievements on various applications ranging from self-driving~\cite{chen2021geosim}, service robots~\cite{gupta2021embodied}, to health and wellbeing~\cite{ravi2016deep}. The model would perform well with the assumption that training and testing data are independent identically distributed (\textit{i.i.d.}); however it seldomly holds for real-world applications. The violation of \textit{i.i.d.} assumption could hinder the performance of deep learning models upon practical deployment. Without loss of generality, we denote the data and label as $X$ and $Y$, the joint distribution as $P(X,Y)$, the latter of which can be formulated by $P(Y)P(X|Y)$. We argue that the reason why current models fail to generalize well in real-world applications is rooted in both the categorical distribution $P(Y)$ and class conditional distribution $P(X|Y)$. 

On one hand, real-world data exhibits {long-tailed distribution} over categories, with only a few classes (\textit{head}) accounting for the major proportions, whilst many more classes (\textit{tail}) presenting extremely limited samples~\cite{zhang2021deep}. For instance, in action recognition, the case of ``\textit{open door}'' is common in daily activities, whereas some actions such as ``\textit{repair door}'' occur much less frequently. This leads to a long-tailed label distribution of $P(Y)$, where conventional training strategies that apply common classification losses (mostly cross-entropy) on instance-level sampled batches would fail. In this case, the trained model would gain high performance on the head classes but behave poorly on tail classes, failing to achieve consistently good performance across all categories. On the other hand, the conditional distribution $P(X|Y)$ is also prone to changes in the real world~\cite{wang2021generalizing}. Different styles of image recognition data, camera viewpoints of action recognition data, acquisition protocols of medical images, etc., would alter the distribution of $P(X|Y)$, leading to diversified distributions, a.k.a. domain shifts.

\begin{figure}[t!]
    \centering
    \includegraphics[width=0.75\linewidth]{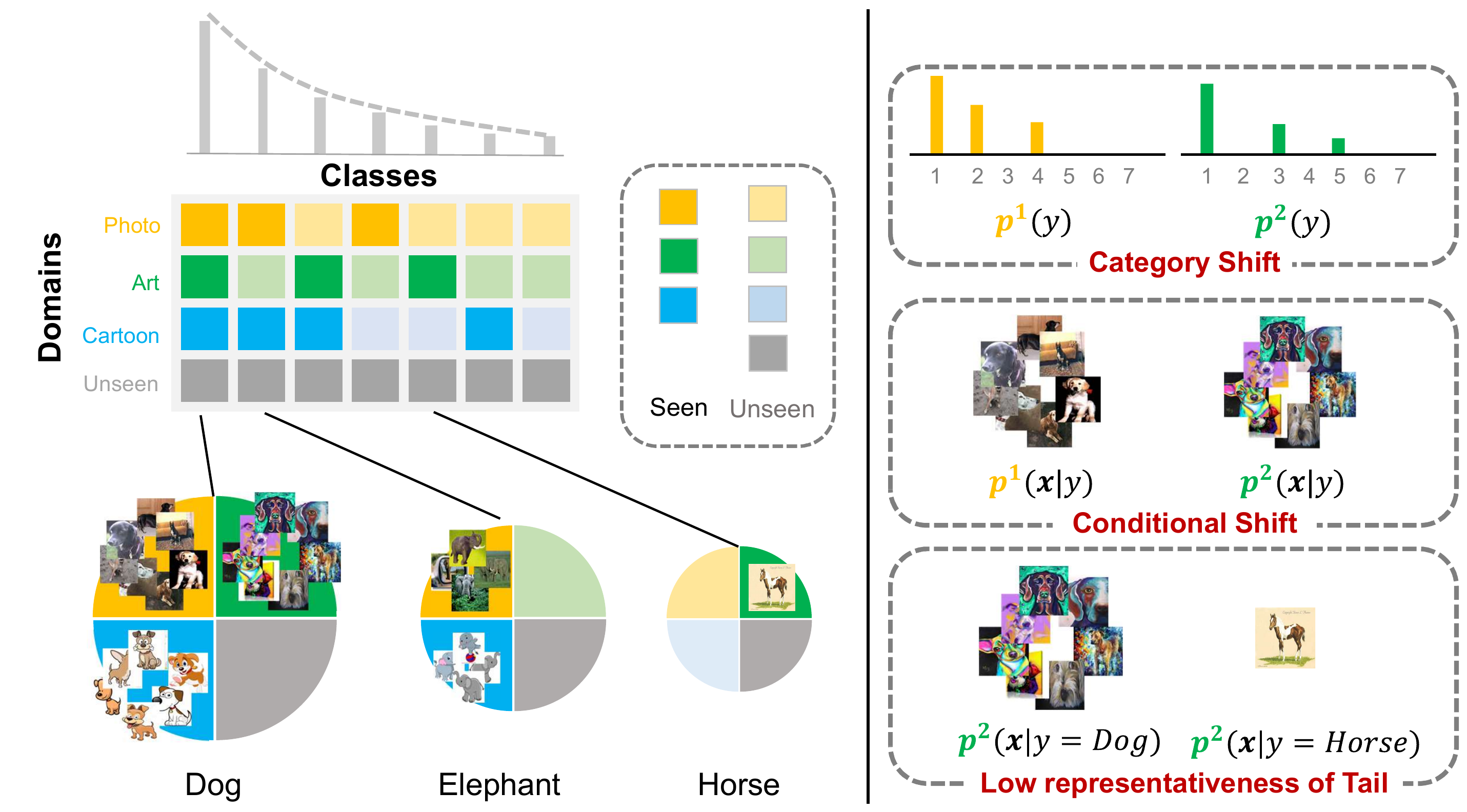}
    \caption{Visual illustrations of the issues complicated with long-tailed category distribution and conditional distribution shifts across domains. The overall dataset\protect\footnotemark[1] is long-tailed distributed over classes. Meanwhile, with data collected from multiple domains (e.g., different styles), head classes (e.g., \textit{Dog}) can be observed in most domains, whereas tail classes (e.g., \textit{Horse}) only contain a few samples in certain specific domains. } 
    \label{fig:fig1}
\end{figure}
 
\footnotetext[1]{Images are adopted from PACS~\cite{li2017deeper}. Its distribution originally is not long-tailed. Here just for intuitive explanation of our focused problem.}

In this regard, long-tailed categorical distribution (\textbf{LT})~\cite{zhang2021deep} and domain shifts (\textbf{DS})~\cite{wang2021generalizing} have been two major issues concerned with real-world datasets. Although increasing research efforts have been made, these two issues are so far tackled individually, with their complex co-existence situation not being considered yet. Existing solutions cannot deal with the entanglement of \textbf{LT} and \textbf{DS}, since a balanced distribution $P(Y)$ or identical $P(X|Y)$ are their prior assumptions of those \textbf{DS} \cite{dou2019domain,wang2021generalizing} and \textbf{LT} \cite{ren2020balanced,jamal2020rethinking,li2021metasaug} solutions, respectively. As we know, in real-world scenarios, these two issues often come together. Take medical image data as an example, the conditional distribution varies across different hospitals, and, there are a large population of patients with common diseases whilst some patients with rare diseases. In addition, the low prevalence of those rare diseases may lead to the 
inclusion of corresponding patients only by certain hospitals. This also similarly applies to many other applications \cite{damen2020rescaling} where the head classes are common in most domains, whilst tail classes only appear in certain domains due to the low-frequency. Such combination of $\lt{}$ and $\ds{}$ leads to a more challenging, yet more practical scenario where $P(Y)$ of each individual domain is not only imbalanced, but also partial. Ideally a reasonable model should be robust across classes and generalize across domains, simultaneously.  

We argue that there are three main challenges posed by the problem of \ltds{} (cf. \figurename~\ref{fig:fig1}). \textbf{1)} Because of the existence of multiple domains, the categorical distributions $P(Y)$ are different across domains. Given the relatively low frequency of non-head classes, their corresponding samples may be collected only in certain domains. As a result, the spurious correlation between non-head classes and domain-specific characteristics might be learned as biased shortcuts. \textbf{2)} The conditional distributions $P(X|Y)$ are varied significantly across domains. It is expected that the model can handle such shifts, with domains aligned and unbiased representations learned. \textbf{3)} It is hard to explicitly model the distribution of tail classes $P(X|Y=tail)$, as only a paucity of domain-specific samples exist. This poses challenges to avoiding overfitting on the tails.   
Hence, research is desperately needed to solve the co-occurrence of these issues in \ltds{}. 

In this work, we propose an effective solution to tackle all of the aforementioned challenges. First, a novel domain-specific distribution calibrated loss is introduced to address the infinite imbalance ratio of each domain. Subsequently, we leverage class distributional embeddings as unbiased semantic features, to align the derived visual representations to unbiased semantic space via the alignment between domain-specific visual prototypes and semantic embeddings. Furthermore, we propose a semantic-similarity guided module by leveraging the knowledge learned from head classes, for implicit augmentation of tail classes.
In addition, to ensure the model is capable of handling out-of-distribution data in unseen domains, a meta-learning framework integrating the above three core modules is proposed to boost the generalization capability.  
To evaluate the effectiveness of the proposed method, we developed two datasets with \ltds{} problems, namely \awalts{} and \imagenetlts{}, and conducted extensive comparison experiments on both datasets. Results demonstrate that our proposed method exceeds state-of-the-art \textbf{LT} or \textbf{DS} methods by a large margin.

\section{Related Works}

\textbf{Long-Tailed Category Distribution.} To learn from class imbalanced training data, one line of existing works aims to manipulate class-wise contributions by resampling~\cite{buda2018systematic,han2005borderline}, reweighting~\cite{huang2016learning,wang2017learning,lin2017focal}, logits adjustment~\cite{ren2020balanced,hong2021disentangling}, and two-stage training~\cite{kang2019decoupling,tang2020long}. Another emerging line has made attempts at ensemble learning under long-tailed settings, such as contrastive learning~\cite{wang2021contrastive}, knowledge distillation~\cite{iscen2021class}, variance-bias calibration~\cite{wang2021longtailed}. In particular, RIDE~\cite{wang2021longtailed} indicates that the predictions of head classes would be of larger intra-class variances, whereas tail classes would exhibit larger biases. This becomes more serious in the \ltds{} scenario, since the intra-class variances are related to not only semantics but also domains shifts; whereas the representations of non-head classes may easily be biased by domain-specific characteristics. Unfortunately, most of the existing methods do not take into account conditional-distribution-shift introduced biases, instead assuming identical in their work. Similar issues exist in recent meta-learning based approaches. To ensure good performance across all classes, recent works investigated the category shift between long-tailed and balanced distributions, and introduced meta-learning strategies to optimize parameters on a held-out balanced meta-test subset~\cite{jamal2020rethinking,li2021metasaug}. This is not applicable for \ltds{}, since it is impossible to sample a held-out meta-test proportion with balanced category distribution without conditional distribution shifts. 

\noindent\textbf{Model Generalization at Domain Shifts.} Domain generalization (DG) aims to develop computational models that are capable of handling data from unseen domains. Existing domain generalization solutions are varied, including aligning intra-class representations across domains \cite{li2018deep}, factoring out domain-specific information \cite{wang2020cross,gu2020cross}, simulating domain gaps via sophisticated training strategies \cite{dou2019domain,li2019episodic}, or performing data augmentation \cite{mancini2020towards,zhou2021mixstyle}. The shortcomings of most solutions become apparent when faced with \ltds{}, as \ltds{} poses imbalanced distribution over a large number of classes. For the methods benefiting from explicit categorical distribution alignment~\cite{li2018deep}, it is computationally prohibitive to design class-specific aligning models, and impossible to align domain-specific tail classes. Furthermore, the large class number makes sampling classes from multiple domains intractable, thus being difficult to cover relatively large portions of the label set in a mini batch \cite{dou2019domain}. Even worse, for those tail classes, since they are only available in certain domains due to the low-frequency, some short-cuts of the classifier may be learned due to the spurious correlation between the domain-specific information and the occurrence of associated tail classes.  

On the other hand, most current domain generalization approaches assume similar categorical distribution across domains, yet this can hardly hold true in the real world~\cite{liu2021domain,shu2021open,you2019universal}. A similar issue has been raised in \cite{shu2021open} referred to as open domain generalization, where the distribution and label sets of each source domain and target domain can be different. Shu~\etal{}~\cite{shu2021open} introduced the domain-specific model in each individual source domain and applied a meta learning strategy to generalize each domain-specific model to other domains, by knowledge distillation from other domains. However, this framework cannot well apply to \ltds{} settings. To be specific, the knowledge derived from each domain is easy to get biased due to the specific long-tailed and incomplete categorical distribution in each domain. Even worse, there is no guarantee that under such bias, the spurious correlation between non-semantic domain-specific characteristics and domain-specific classes can be avoided. 

\noindent\textbf{Cross-Modal Representation Learning.}
Leveraging information from multiple modalities is popular for related multi-modality applications~\cite{socher2013zero,radford2021learning} to facilitate effective representation learning of each individual modality. One of the related applications is few/zero-shot learning, where one line of research aims to establish the relationship between semantic space and visual space~\cite{xian2019f}. Maniyar~\etal{}~\cite{maniyar2020zero} leveraged the semantic space to enable zero-shot domain generalization, and Mancini~\etal{}~\cite{mancini2020towards} used the semantic embeddings as the classifier for zero-shot learning in unseen domains. These inspired our work; however as a different task setting, our goal is to derive unbiased predictor under long-tailed settings such that the missing classes in seen domains can be recognized as well.

Recently, Samuel~\etal{}~\cite{samuel2021generalized} leveraged class descriptors to facilitate long-tailed classification. It developed a dual network to derive both visual features and semantic features from the input image, and then fused these two together to boost the performance of long-tailed classification. Although it applied semantic embeddings similar to our work, our task aims to address a more challenging problem, where both imbalance and conditional distribution shifts exist.

\section{Methodologies}
\subsection{Problem Setup and Preliminaries}
\begin{figure}
    \centering
    \includegraphics[width=0.7\linewidth]{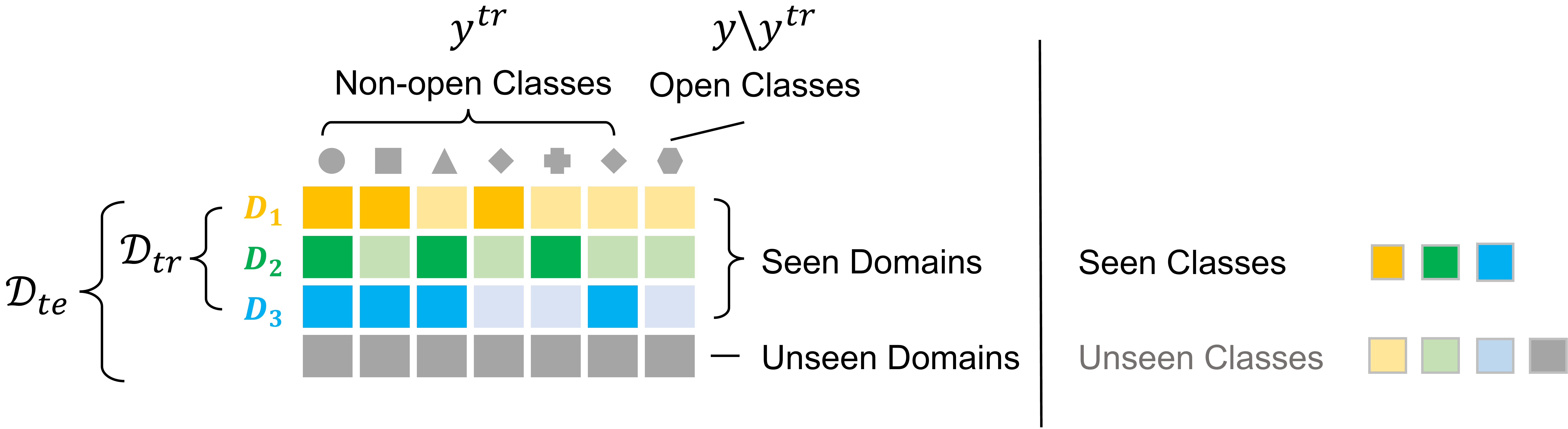}
    \caption{Illustrations of domain splits, open/non-open classes, and seen/unseen classes.} 
    \label{fig:notation}
\end{figure}
We denote the input and label spaces as $\mathcal{X}$ and $\mathcal{Y}$, and the domain space as $\mathcal{D}$. $\mathcal{D}$ consists of totally $K$ domains $\{D_k\}_{k=1}^K$ and there are totally $C$ categories in the label space. Each sample is denoted as $\{\bx{}_i,{y_i}, {d_i}\}$, where $i$ indicates the sample index, $\bx{}_i$ the input sample, $y_i$ the ground truth label, and $d_i$ the domain index; $1\leq d_i \leq K$. The training domains and testing domains are denoted as $\mathcal{D}_{tr}$ and $\mathcal{D}_{te}$, respectively, where $\mathcal{D}_{tr}\subset\mathcal{D}$ and $\mathcal{D}_{te}=\mathcal{D}$. The categorical distribution $p^k(y)$ of each training domain $k$ follows a long-tailed distribution, and the low prevalence of tail classes may lead to the failure of collecting training samples from rare classes, i.e., $\mathcal{Y}^k \subset \mathcal{Y}$. We denote the label set of all training data as $\mathcal{Y}^{tr}=\bigcup_{k=1}^{|\mathcal{D}_{tr}|}\mathcal{Y}^k$. To test the overall performance across all classes, test data is sampled under balanced distribution over all classes. Since there might be domain-specific non-head classes in each domain, open classes exist in the testing domains, namely $\mathcal{Y}^{tr} \subset \mathcal{Y}$, $\mathcal{Y}^{te} = \mathcal{Y}$. A visual illustration is presented in \figurename~\ref{fig:notation}. Our ultimate goal is to build a computational model that is able to recognize all the non-open classes across domains, as well as open classes belonging to  $\mathcal{Y}\setminus\mathcal{Y}^{tr}$. 

The computational model $g: \mathcal{X}\rightarrow\mathcal{Y}$ maps raw input to the final prediction. Following previous domain generalization works~\cite{dou2019domain}, it can be decoupled into a feature extractor $f$ and a head classifier $h$, where $f:\mathcal{X}\rightarrow \mathcal{Z}$, and $h:\mathcal{Z}\rightarrow\mathbb{R}^{C}$. The final prediction $\hat{y}=g(\bx) = h\circ f(\bx)$. With the loss function denoted as $\mathcal{L}(h \circ f(\bx),y)$, we derive the estimated error $\epsilon$ on test data as:

\begin{equation}
\begin{aligned}
    \epsilon &= \mathbb{E}_{m\sim P_{\mathcal{D}_{te}}}\mathbb{E}_{(\bx,y)\sim p^m(\bx,y)} \mathcal{L}(h\circ f(\bx), y) \\
        &= \mathbb{E}_{\begin{subarray}{l}{m\sim P_{\mathcal{D}_{te}}}\\ {n\sim P_{\mathcal{D}_{tr}} }\end{subarray}} \mathbb{E}_{(\bx,y)\sim p^n(\bx,y)} \mathcal{L}(h\circ f(\bx), y) \frac{p^m(f(\bx),y)}{p^n(f(\bx),y)}  \\
          &= \mathbb{E}_{\begin{subarray}{l}{m\sim P_{\mathcal{D}_{te}}}\\ {n\sim P_{\mathcal{D}_{tr}} }\end{subarray}} \mathbb{E}_{(\bx,y)\sim p^n(\bx,y)} \mathcal{L}(h\circ f(\bx), y) \frac{p^m(y)p^m(f(\bx)|y)}{p^n(y)p^n(f(\bx)|y)},
\end{aligned}
\label{eq:erm}
\end{equation}

where $P_{\mathcal{D}_{tr}} \& P_{\mathcal{D}_{te}}$ denotes the probability of sampling data from training or testing domains. 

To minimize $\epsilon$ as in Equation \eqref{eq:erm}, it is of paramount significance to model the term $ \frac{p^{te}(y)p^{te}(f(\bx)|y)}{p^{tr}(y)p^{tr}(f(\bx)|y)}$ to ensure the robustness under \ltds{}. However, there exist several issues that are challenging to resolve:

\textbf{1)} $p^n(y)$ of each individual training domain is imbalanced. Even through a balanced resampling on seen classes, those unseen classes of each individual domain still leads to ``imbalance'' with an infinite ratio. 

\textbf{2)} It comes with challenges in aligning the distribution of $p^n(f(\bx)|y)$ to an unbiased and semantically meaningful space, since there are some classes unseen in each individual domain, especially for those tail classes.   

\textbf{3)} The distributions of tail classes $p(f(\bx)|y)$ are difficult to model compared to head classes, as caused by the limited sample number in certain domains. 

\textbf{4)} Since we aim to model and align $p^n(f(\bx)|y)$ rather than $p^n(\bx|y)$, it is important to make sure that $f$ is able to handle out-of-distribution data itself, thus enabling extracting domain-invariant and discriminative features by $f(\bx)$. 

It should be noted that Equation \eqref{eq:erm} gets some inspirations from the recent work~\cite{jamal2020rethinking}, while they are conceptually different. To be specific, Jamal~\etal{}~\cite{jamal2020rethinking} considers the distribution shifts across long-tailed and balanced distributions; whereas the shifts of $P(X|Y)$ like style changes are not taken into consideration.   

In the following, we first go through the core functional blocks to address aforementioned issues, followed by a meta-learning based framework to integrate these functional blocks.

\subsection{Core Functional Blocks}
\paragraph{\textbf{Distribution Calibrated Classification Loss}-Model $\frac{p^m(y)}{p^n(y)}$.}
Considering the term $\frac{p^m(y)}{p^n(y)}$, we aim to tackle the imbalance of training data $p^n(y)$ so as to work on balanced distribution $p^m(y)$ with distribution calibrated classification loss. 
In \cite{ren2020balanced}, Ren~\etal{} proposed a variant version of softmax function to approximate the discrepancy of the posterior distributions between training and testing data. Similar ideas have also been introduced in~\cite{hong2021disentangling}. 
Based on \cite{ren2020balanced,hong2021disentangling}, the distribution calibrated classification loss is formulated as:  
\begin{equation}
    \mathcal{L}_{dc}(\bx_i, y_i, d_i; f, h) = -\log\frac{n_{y_i}^{d_i} \exp{([h\circ f(\bx_i)]_{y_i}})}{\sum_{c=1}^C n_c^{d_i} \exp({[h\circ f(\bx_i)]_{c})}},
   \label{eq:dc}
\end{equation}

\noindent where $n_c^{d_i}$ denotes the sample number of class $c$ in the training domain $D_{d_i}$. Please see supplementary material for proof.  

\begin{figure}[btp!]
\begin{minipage}[b]{\linewidth}
   \centering
    \includegraphics[width=0.65\linewidth]{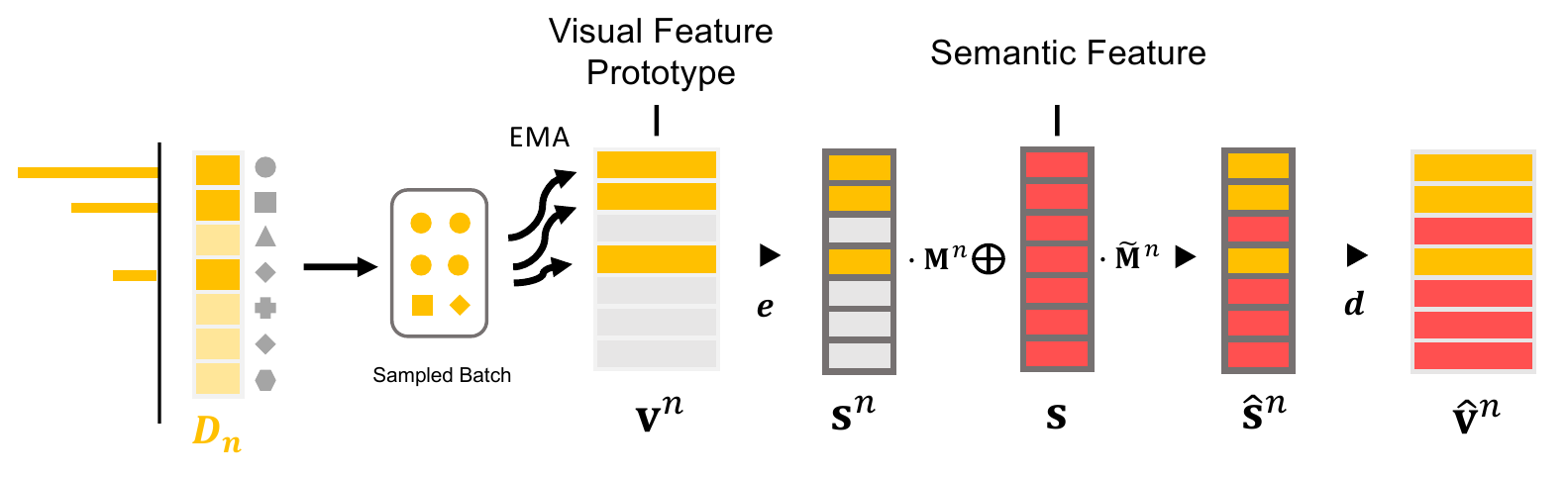}
    \caption{Transformation between \textbf{semantic space} $\mathbf{s}$ and \textbf{visual feature prototypes} $\mathbf{v}^n$ of each domain $D_n$. $\mathbf{s}$ denotes semantic features based on word embeddings from class names or sentence embeddings from class descriptors. In each iteration, based on sampled batch from domain $D_n$, its visual prototype $\mathbf{v}^n$ is updated by exponential moving average (EMA). After transforming it to the semantic space $\mathbf{s}^n$ by $e$, the missing entries in $\mathbf{v}^n$ are filled by the corresponding semantic embeddings in $\mathbf{s}$ to derive the complete $\hat{\mathbf{s}}^n$ and then converted back to the visual space as $\hat{\mathbf{v}}^n$.} 
    \label{fig:v2s.}
\end{minipage}
\begin{minipage}[b]{0.2\linewidth}
\vfill\includegraphics[width=0.75\linewidth]{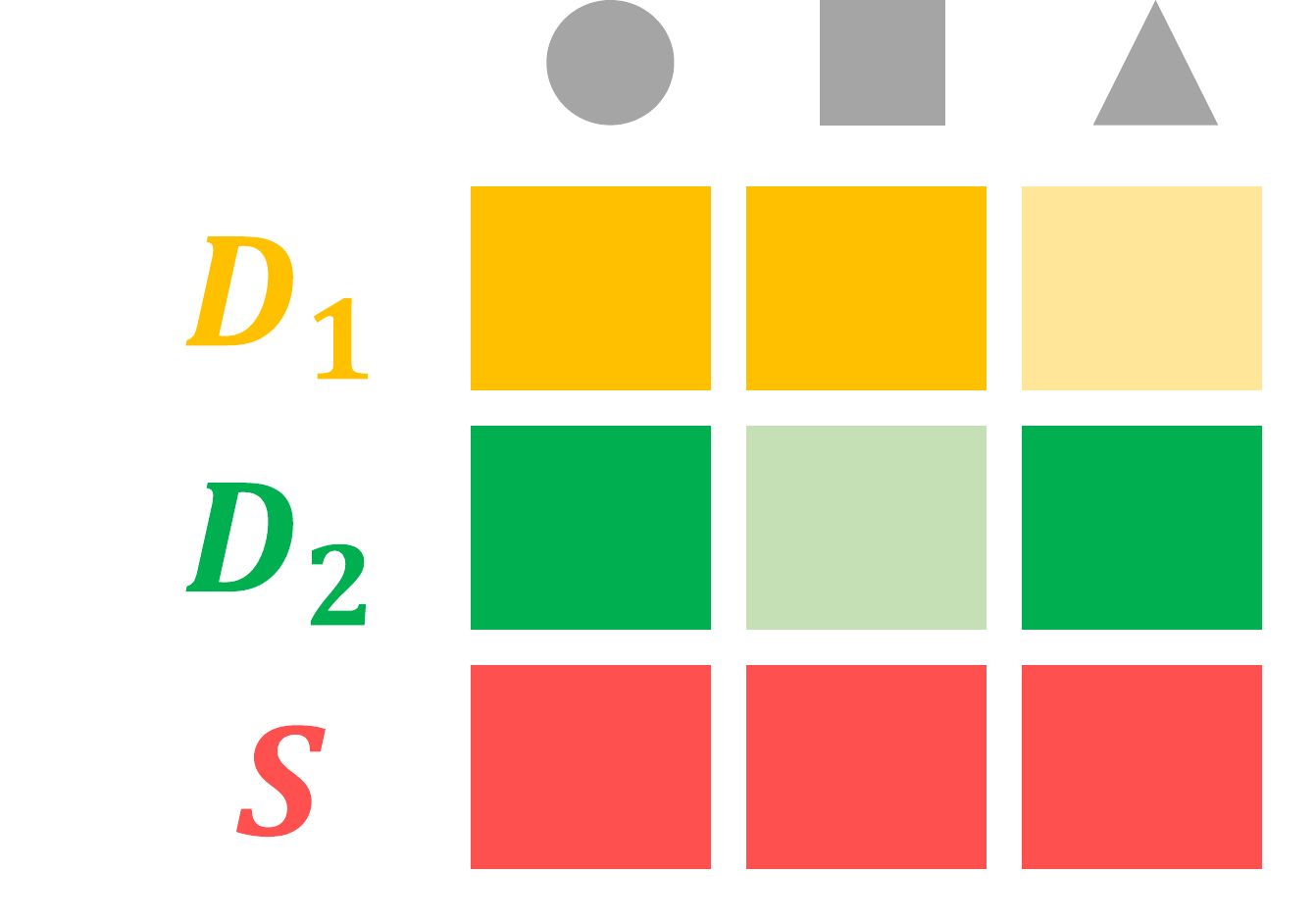}
\includegraphics[width=\linewidth]{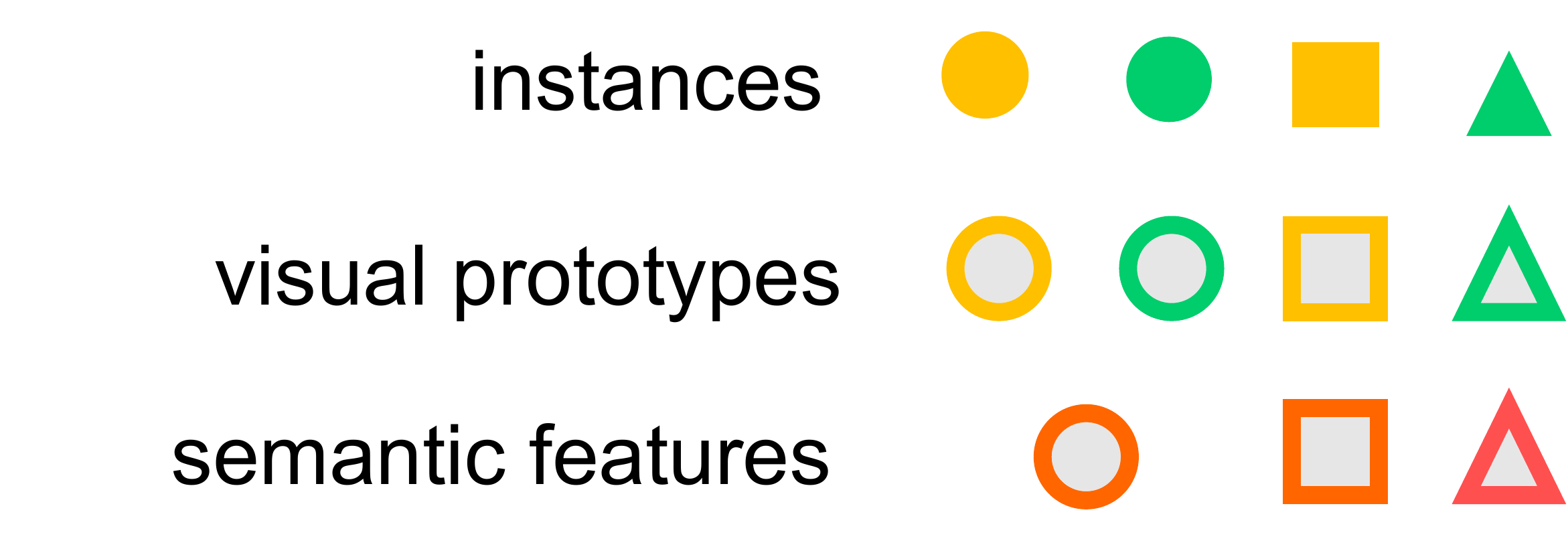}
\end{minipage}
\hfill
\begin{minipage}[b]{0.35\linewidth}
\includegraphics[width=0.85\linewidth]{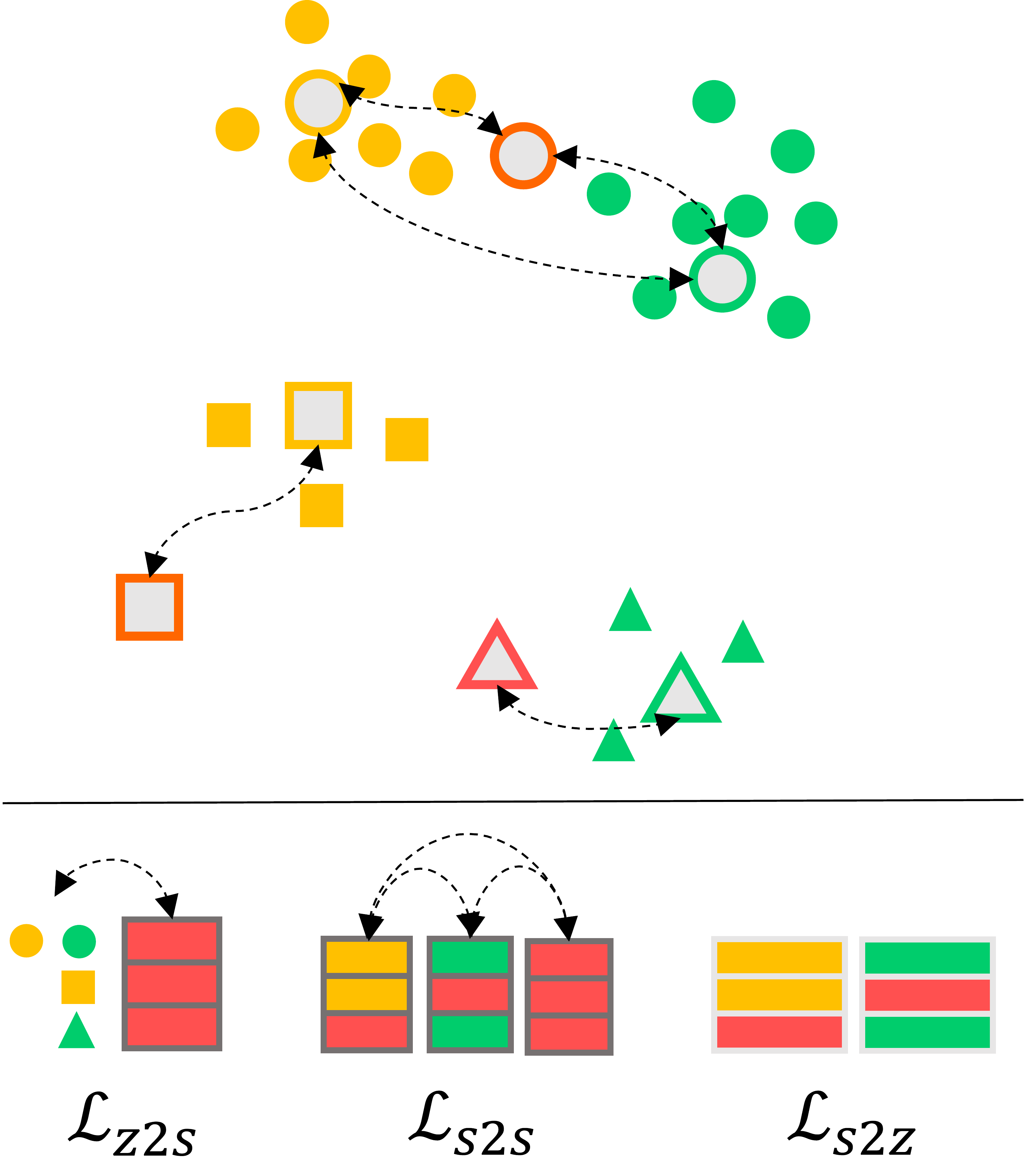}
\caption{Illustration of visual-semantic mapping to derive domain-invariant and unbiased $p^n(f(\bx)|y)$ across domains.}
\label{fig:v2salign}
\end{minipage}
\hfill
\begin{minipage}[b]{0.4\linewidth}
\includegraphics[width=0.92\linewidth]{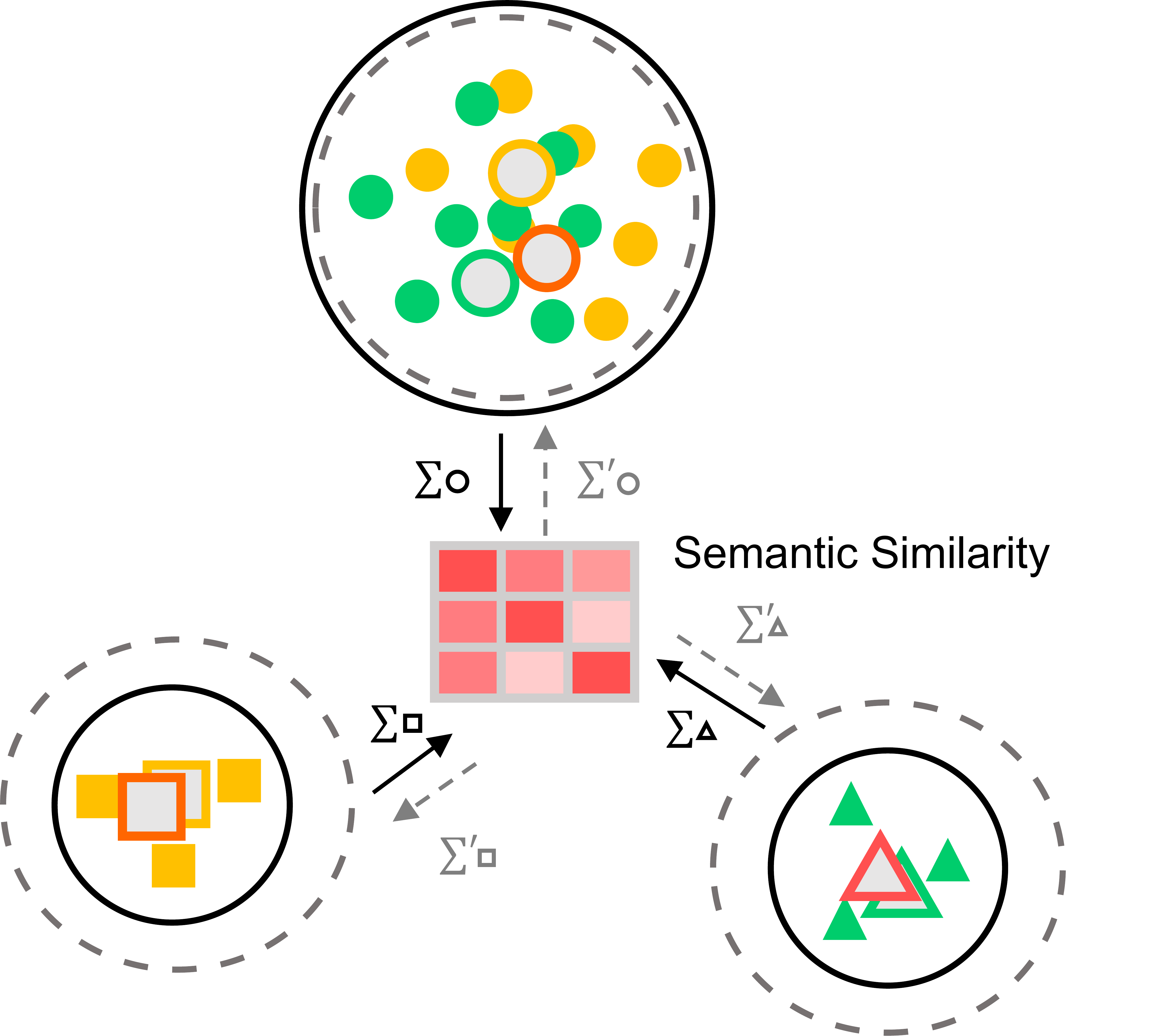}
\caption{Illustration of semantic-similarity guided augmentation to facilitate distribution modelling of tail classes by utilizing the semantic relationship between tail and head classes.}
\label{fig:aug}
\end{minipage}
\end{figure}

\paragraph{\textbf{Visual-Semantic Mapping}-Align $p^n(f(\bx)|y)$.}
To ensure unbiased and semantically meaningful representations, we leverage the semantic embeddings based on word embeddings from class names or sentence embeddings from class descriptors, inspired by existing zero-shot learning works~\cite{maniyar2020zero,mancini2020towards}. With totally $C$ classes and feature dim as $d_s$, the semantic embedding is denoted as $\mathbf{s}\in\mathbb{R}^{C\times d_s}$, with its $c$ element $\mathbf{s}_c$ corresponding to the embedding of class $c$. 

On the other hand, for each domain $n$ with classes $C$ in total and feature dim as $d_v$, we have a visual prototype $\mathbf{v}^n\in\mathbb{R}^{C\times d_v}$,  which is derived in an online manner by exponential moving average (EMA). Some entries of $\mathbf{v}^n$ are probably empty caused by the missing categories in each individual domain $n$. The index mask of valid entries is denoted as $\mathbf{M}^n$ for convenience. 

To achieve the alignment between visual feature prototype \{$\mathbf{v}^n$\} and $\mathbf{s}$, we introduce another two functions $e$ and $d$, where $e:\mathcal{Z}\rightarrow\mathcal{S}$ and $d: \mathcal{S}\rightarrow\mathcal{Z}$. \{$\mathbf{v}^n$\} is firstly transformed to the semantic space by $e$ as $\mathbf{s}^n$ and subsequently, the missing entries of $\mathbf{s}^n$ are filled by the corresponding semantic features in $\mathbf{s}$ to $\hat{\mathbf{s}}^n$, by $e(\mathbf{v}^n)\cdot\mathbf{M}^n\oplus\mathbf{s}\cdot\tilde{\mathbf{M}}^n$. By the function $d$, $\hat{\mathbf{s}}^n$ is transformed back to visual space as $\hat{\mathbf{v}}^n$. The data flow is visualized in \figurename~\ref{fig:v2s.}. 

We introduce three typical losses to fulfill the goal of visual-semantic alignment, which are visually illustrated in \figurename~\ref{fig:v2salign}. First of all, $\mathcal{L}_{z2s}$ of Equation~\eqref{eq:z2s} is utilized to align each training sample to its corresponding semantic feature. In Equation~\eqref{eq:z2s}, we adopt margin contrastive loss on a unit-normalized embedding space, where the margin $\alpha$ is inspired by~\cite{wang2018additive} as to encourage more tight distribution of intra-class embeddings, and $\tau$ is the temperature constant.  

Furthermore, to avoid shifts across domains, we enforce a cross-prototype contrastive loss as in Equation~\eqref{eq:s2s}, which aims to decrease intra-class inter-domain discrepancies and enlarge inter-class distances. Additionally, to enforce the manifold constraint of our learned feature representation, we convert $\hat{\mathbf{s}}^n_v$ back to visual space $\hat{\mathbf{v}}^n$ and apply $\mathcal{L}_{s2s}$ as in Equation~\eqref{eq:s2z}. It applied the classification loss based on the visual classifier $h$, and also aims to align $\hat{\mathbf{v}}^n$ to semantic embeddings by $\mathcal{L}_{s2s}([e(\hat{\mathbf{v}}^n)], \mathbf{s})$ similar to a cycle loss~\cite{zhu2017unpaired}. To deal with the class imbalance in each batch, $\mathcal{L}_{z2s}$ is further integrated with the last module when calculating the loss. 

\begin{figure}
\begin{minipage}[b]{\linewidth}
 \begin{equation}
\mathcal{L}_{z2s}(\bx_i,\! y_i,\!\mathbf{s};e,\!f)\!=\!-\!\log\frac{\exp(([e\circ f(\bx_i)]^\intercal \mathbf{s}_{y_i}-\alpha)/{\tau})}{\exp(([e\!\circ\! f(\bx_i)]^\intercal \mathbf{s}_{y_i}\!\!-\!\alpha)/{\tau}) \!+\!\!\underset{{j\neq y_i}}{\sum}\!\exp(([e\!\circ\! f(\bx_i)]^\intercal \mathbf{s}_{j})/{\tau})}, \label{eq:z2s}
 \end{equation}
\begin{equation}
 \mathcal{L}_{s2s}(\hat{\mathbf{s}}^m, \hat{\mathbf{s}}^n)\!=\! \mathbb{E}_c \Big [ -\!\log\frac{\exp(({\hat{\mathbf{s}}^{m\intercal}_c} \hat{\mathbf{s}}^n_{c}\!-\!\alpha)/{\tau})}{\exp(({\hat{\mathbf{s}}^{m\intercal}_c} \hat{\mathbf{s}}^n_{c}\!-\!\alpha)/{\tau}) \!+\underbrace{\!\underset{{j\neq c}}{\sum}\exp({\hat{\mathbf{s}}^{m\intercal}_c} \hat{\mathbf{s}}^n_{j}/\tau)\!+\!\underset{{j\neq c}}{\sum}\exp({\hat{\mathbf{s}}^{m\intercal}_c} {\hat{\mathbf{s}}^m_{j}}/\tau)}_{\text{enlarge inter-class inter-/intra-domain distance}}}\Big ],
 \label{eq:s2s}
\end{equation}
\begin{equation}
 \mathcal{L}_{s2z}({\hat{\mathbf{v}}}^n; e, h) = \mathbb{E}_{i} \Big [-\log \frac{\exp([h({\hat{\mathbf{v}}}_{i}^n)]_{i})}{\sum_{c=1}^C\exp([h({\hat{\mathbf{v}}^n}_{i})]_c)} \Big ]  + \mathcal{L}_{s2s}([e(\hat{\mathbf{v}}^n)], \mathbf{s}).
\label{eq:s2z}
\end{equation}
\end{minipage}
\end{figure}

\paragraph{\textbf{Semantic-Similarity Guided Augmentation}-Model $p(f(\bx)|y)$.}
Another troubling issue lies in the poor diversity of tail classes. In addition to achieving semantically meaningful and unbiased representations, it is also expected that overfitting on the tail classes can be avoided. It emphasizes the importance of adding to the diversity and richness of tail classes. Therefore, a semantic similarity guided feature augmentation method is proposed as below.  

We define the conditional feature distribution (assumed as multi-variate Gaussian distribution, aggregated from all domains) as $p(f(\bx)|c)\sim\mathcal{N}(\boldsymbol{\mu}_c, \mathbf{\Sigma}_c)$. The classifier $h$ is composed of a weight matrix $[{w}_1,..., {w}_C]^\intercal$ and biases $[b_1,..., b_C]^\intercal$. Without loss of generality and for simplicity, we only consider the weight matrix in the following. The upper bound of softmax cross entropy loss \cite{wang2019implicit} can therefore be derived as in Equation~\eqref{eq:surro}, with proof in supplementary material.

\begin{equation}
\begin{aligned}
    & \mathbb{E}_{f(\bx_i)} \Big [-\log\frac{\exp(w_{y_i}^\intercal f(\bx_i))}{\sum_{c=1}^C \exp(w_c^\intercal f(\bx_i))} \Big ] \\
     & \leqslant \log \Big [\sum_{c=1}^C \exp((w_c^\intercal-w_{y_i}^\intercal)\mathbf{\boldsymbol{\mu}}_{y_i} + \frac{\lambda}{2}(w_c^\intercal-w_{y_i}^\intercal) \mathbf{\Sigma}_{y_i} (w_c - w_{y_i}))\Big ].
\end{aligned} 
\label{eq:surro}
\end{equation}

This indicates that by adding the penalty of $\frac{\lambda}{2}(w_c^\intercal-w_{y_i}^\intercal) \mathbf{\Sigma}_{y_i} (w_c - w_{y_i})$, the up-boundary of classification loss can be approximated by implicit augmentation, where $\lambda$ can be considered as a term to control the augmentation degree \cite{wang2019implicit}. 

Thus far, we assume a nearly-identical visual space (i.e., similar $f(\bx|y=c)\approx\boldsymbol{\mu}_c$ across domains) after visual-semantic mapping; however the estimation of $\mathbf{\Sigma}_c$ is hardly possible for tail classes. 
Guided by the semantic inter-class relationship from $\mathbf{s}$, we select the top $k$ classes that are most similar to the corresponding class $c$ (including $c$ itself). This stems from the observation that similar classes are supposed to have similar semantic variances. For example, \textit{deer} and \textit{antelope} may share similar characteristics of the variations of shape, color, etc. Motivated by this, we introduce a weighted covariance estimation strategy to leverage the knowledge learned from head classes, 

\begin{equation}
\centering
\begin{aligned}
        Sim_c = \{\mathbf{s}_c^\intercal\mathbf{s}_i|i=1,2,&...,C\} ; \mathbf{k}_c = \{i | \mathbf{s}_c^\intercal\mathbf{s}_i \in topk(Sim_c)\}, \\
   \mathbf{\Sigma}'_c &= \frac{\sum_{k\in \mathbf{k}_c}{n_k}\mathbf{\Sigma}_k}{\sum_{k\in \mathbf{k}_c} n_k}. 
   \label{eq:updateaug}
\end{aligned}
\end{equation}

Afterwards, we applied an surrogate loss introduced in Equation \eqref{eq:surro} to optimize the boundary of classification loss by adding implicit augmentation terms:
\begin{equation}
    \mathcal{L}_{aug}(\bx_i, y_i; f,h) = -\!\log \frac{\exp(w_{y_i}^\intercal f(\bx_i))}{\sum_{c=1}^C \exp(w_{y_i}^\intercal f(\bx_i)+\frac{\lambda}{2}(w_c-w_{y_i})^\intercal \mathbf{\Sigma}'_{y_i}(w_c-w_{y_i}))}.
    \label{eq:aug}
\end{equation}
In practice, the covariance $\mathbf{\Sigma}$ is online calculated from $T_{\Sigma}$ steps onwards to avoid the effect of inter-domain variances.   

\subsection{Meta-Learning Based Generalization\\}

The objective of meta-learning is to ensure that the trained models are robust against domain shifts and perform well on all seen and unseen classes. If combining the functional blocks introduced above, we can obtain relatively good results by conventional training strategies. However, the generalization capability on unforeseen domains is not guaranteed. Thus, we apply meta-learning to simulate the domain distribution gaps in an episodic manner, inspired by previous DG works~\cite{dou2019domain}. The optimization process is illustrated in Algorithm~\ref{alg}. For each iteration, the training domains are randomly divided into two splits, $\mathcal{D}_{mtr}$ and $\mathcal{D}_{mte}$. We make sure that data in $\mathcal{D}_{mte}$ always come from different domains from $\mathcal{D}_{mtr}$. After training on $\mathcal{D}_{mtr}$, the model is expected to perform well on unseen domains (especially for domain-unique tail classes) in $\mathcal{D}_{mte}$. 

\paragraph{\textbf{Meta Train.}} Over the course of meta-training, we make the model trained on $\mathcal{D}_{mtr}$ able to derive semantically meaningful and unbiased representations. With a batch data of size $B$ sampled from each domain in $\mathcal{D}_{mtr}$, i.e., $\{\bx_{i}, y_i, d_i\}_{i=1}^{B\times|\mathcal{D}_{mtr}|}$, we exert the following typical loss functions. 

First of all, in order to calibrate the loss from imbalanced distributions to balanced ones, we apply the domain calibrated softmax loss with $\mathcal{L}_{Cls} = \frac{1}{{B\times|\mathcal{D}_{mtr}|}} \sum_{i}\mathcal{L}_{dc}(\bx_{i}, y_i, d_i; f, h)$. In this way, we can improve the performance over all classes and not propagating discouraging gradients to unseen classes. Subsequently, to build unbiased representations, the visual-semantic alignment loss $\mathcal{L}_{Z2S} = \frac{1}{{B\times|\mathcal{D}_{mtr}|}} \sum_{i}\mathcal{L}_{z2s}(f(\bx_i), y_i, \mathbf{s}; e)$ is adopted to align the embeddings in the semantic space. Furthermore, to enable domain-invariant feature learning as well as to avoid the prohibitive costs of sampling all classes in all domains when the class number is huge, a prototype alignment loss is utilized, $\mathcal{L}_{S2S} = \mathbb{E}_{m,n\in{\mathcal{D}_{mtr}}} \mathcal{L}_{s2s}(\hat{\mathbf{s}}^m, \hat{\mathbf{s}}^n) +  \mathbb{E}_{n\in{\mathcal{D}_{mtr}}} \mathcal{L}_{s2s}(\hat{\mathbf{s}}^n,\mathbf{s})$. 
In addition, we apply $\mathcal{L}_{S2Z}=\mathbb{E}_{n\in \mathcal{D}_{mtr}}\mathcal{L}_{z2s}({\hat{\mathbf{v}}}^n;e,h)$ to further constrain the semantic manifold. 

With intra-class inter-domain distribution aligned, the intra-class variances would be more semantically relevant. We then track the covariance of $f(\bx)$ feature distribution from $T_\Sigma$ steps onwards, and apply the surrogate augmentation loss, $\mathcal{L}_{Aug} = \frac{1}{{B\times|\mathcal{D}_{mtr}|}} \sum_{i}\mathcal{L}_{aug}(\bx_i, y_i; f,h)$, to increase the diversity of feature distributions, especially for the tail classes. $\mathcal{L}_{Aug}$ is set to $0$ before $T_\Sigma$.  

The overall meta-training loss is formulated as $\mathcal{L}_{mtr}=\mathcal{L}_{Cls} + w_1 \mathcal{L}_{Z2S} + w_2 \mathcal{L}_{S2S} + w_3 \mathcal{L}_{S2Z} + w_4 \mathcal{L}_{Aug}$, where $w_1$, $w_2$, $w_3$, $w_4$ are weight hyperparameters. The model parameters $\theta\{f,h,e,d\}$ at step $t$ are firstly updated based on $\mathcal{L}_{mtr}$ with an optimization step with learning rate $\beta_1$:
\begin{equation}
    \theta^t\{f',h',e',d'\} = \theta^t\{f,h,e,d\}-\beta_1 \triangledown \mathcal{L}_{mtr}.
\end{equation}

\begin{algorithm}[t]
\begin{minipage}{\linewidth}
\small{
\caption{Meta-learning for long-tailed domain generalization.}\label{alg:cap}
\label{alg}
\hspace*{0.02in} {\bf Input:} 
Training set $\mathcal{D}^{tr}$; semantic embeddings $\mathbf{s}$; initialized visual prototype ${\mathbf{v}^n}$. \\
\hspace*{0.02in} {\bf Hyperparameters:} Steps $T_{\Sigma},T_{max}$; Weights $w_1, w_2, w_3, w_4, w_{mte}$; LR $\beta_1, \beta_2$.\\ 
\hspace*{0.02in} {\bf Initialized model parameters:} Feature extractor $f$; classifier $h$; models $e$ and $d$. \\
\hspace*{0.02in} {\bf Output:} 
Optimized models $f$ and $h$.
\begin{algorithmic}[1]
\For{$t$ $\leq$ $T_{max}$}
\State Randomly split $\mathcal{D}_{tr}$ into $\mathcal{D}_{\textcolor{red}{mtr}}$ and $\mathcal{D}_{\textcolor{blue}{mte}}$. 
\State
\State Sample $\{\mathbf{x}_{i}, y_i, d_i\}_{i=1}^{B\times|\mathcal{D}_{mtr}|}$ from $\mathcal{D}_{\textcolor{red}{mtr}}$. \Comment{\small{Meta Train.}}
\State Calculate losses: $\mathcal{L}_{Cls}$, $\mathcal{L}_{Z2S}$.
\State \textit{Update} $\{\mathbf{v}^n|n \in \mathcal{D}_{mtr}\}$ based on $f(\mathbf{x}_i)$. 
\State Calculate new $\{\hat{\mathbf{s}}^n, \hat{\mathbf{v}}^n | n\in \mathcal{D}_{mtr}\}$ based on $e, d, \mathbf{v}$. 
\State \textit{Update} $\mathbf{\Sigma}$ $\mathbf{\Sigma}'$ when {$t$ $\geq$ $T_\Sigma$}.  
\State Calculate losses: $\mathcal{L}_{S2S}$, $\mathcal{L}_{S2Z}$, and $\mathcal{L}_{Aug}$.
\State Calculate meta-training loss: $\mathcal{L}_{mtr}\!=\!\mathcal{L}_{Cls}\!+\!w_1 \mathcal{L}_{Z2S}\!+\!w_2 \mathcal{L}_{S2S}\!+\!w_3 \mathcal{L}_{S2Z}\!+\!w_4 \mathcal{L}_{Aug}$.
\State Calculate new $\theta^t\{f',h',e',d'\} = \theta^t\{f,h,e,d\}-\beta_1 \triangledown \mathcal{L}_{mtr}$.
\State
\State Sample $\{\mathbf{x}_{i}, y_i, d_i\}_{i=1}^{B\times|\mathcal{D}_{mte}|}$ from $\mathcal{D}_{\textcolor{blue}{mte}}$. \Comment{\small{Meta Test.}}
\State Calculate losses: $\mathcal{L}_{MCls}$, $\mathcal{L}_{MZ2S}$, $\mathcal{L}_{MAug}$.
\State Calculate new $\{\hat{\mathbf{s}}^{n\prime}, \hat{\mathbf{v}}^{n\prime} | n\in \mathcal{D}_{\textcolor{red}{mtr}}\}$ based on $e', d', \{\mathbf{v}^n$\}.
\State Calculate meta-testing loss: $\mathcal{L}_{mte} =\mathcal{L}_{MCls} + w_1 \mathcal{L}_{MZ2S} + w_4 \mathcal{L}_{MAug}$.
\State \textit{Update} $\theta^{t+1}\{f,h,e,d\} = \theta^t\{f,h,e,d\} - \beta_2\triangledown(\mathcal{L}_{mtr}+w_{mte}\mathcal{L}_{mte})$.
\EndFor
\end{algorithmic}
}
\end{minipage}
\end{algorithm}

\paragraph{\textbf{Meta Test.}} The model optimized on $\mathcal{D}_{mtr}$ is expected to perform well on held-out domains $\mathcal{D}_{mte}$. In other words, the optimized representations $\theta\{f',h',e',d'\}$ should be unbiased and semantically meaningful, generalizing well when faced with label distribution shifts and novel domain-specific classes. With samples $\{\bx_{i}, y_i, d_i\}_{i=1}^{B\times|\mathcal{D}_{mte}|}$ from $\mathcal{D}_{mte}$, the following losses are utilized.
First comes the calibrated classification loss $\mathcal{L}_{MCls}=\frac{1}{{B\times|\mathcal{D}_{mte}|}} \sum_i \mathcal{L}_{dc}(\bx_{i}, y_i, d_i; f', h')$. Subsequently, with $\hat{\mathbf{s}}^n$ of meta training domains updated to $\hat{\mathbf{s}}^{n\prime}$, another loss $\mathcal{L}_{MZ2S} = \frac{1}{{B\times|\mathcal{D}_{mte}|}} \sum_{i}\mathcal{L}_{z2s}(f'(\bx_i), y_i, \mathbf{s}; e')+ \mathbb{E}_{n\in{\mathcal{D}_{mtr}}}\mathcal{L}_{z2s}(f'(\bx_i), y_i, \hat{\mathbf{s}}^{n\prime}; e')$ aims to align the visual features to both the semantic embeddings $\mathbf{s}$ and the visual prototypes of $\mathcal{D}_{mtr}$. This ensures that the knowledge extracted from meta-training steps are domain-invariant and semantically meaningful across all classes seen in $\mathcal{D}_{mte}$. In addition, the surrogate augmentation loss is enforced from $T_\Sigma$ onwards, to increase the feature diversity, $\mathcal{L}_{MAug} = \frac{1}{{B\times|\mathcal{D}_{mte}|}} \sum\mathcal{L}_{aug}(\bx_i, y_i; f',h')$.  

The overall meta-test loss is $\mathcal{L}_{mte} =\mathcal{L}_{MCls} + w_1 \mathcal{L}_{MZ2S} + w_4 \mathcal{L}_{MAug}$, and we finally update the $\theta\{f,h,e,d\}$ based on $\mathcal{L}_{mtr}+w_{mte}\mathcal{L}_{mte}$ by learning rate $\beta_2$:

\begin{equation}
\theta^{t+1}\{f,h,e,d\} = \theta^t\{f,h,e,d\} - \beta_2\triangledown(\mathcal{L}_{mtr}+w_{mte}\mathcal{L}_{mte}).
\end{equation}

\section{Experiments}
\subsection{Experimental Settings}
\subsubsection{Datasets.}
\begin{figure}[btp!]
\begin{minipage}[b!]{0.5\linewidth}
\centering
\includegraphics[width=\linewidth]{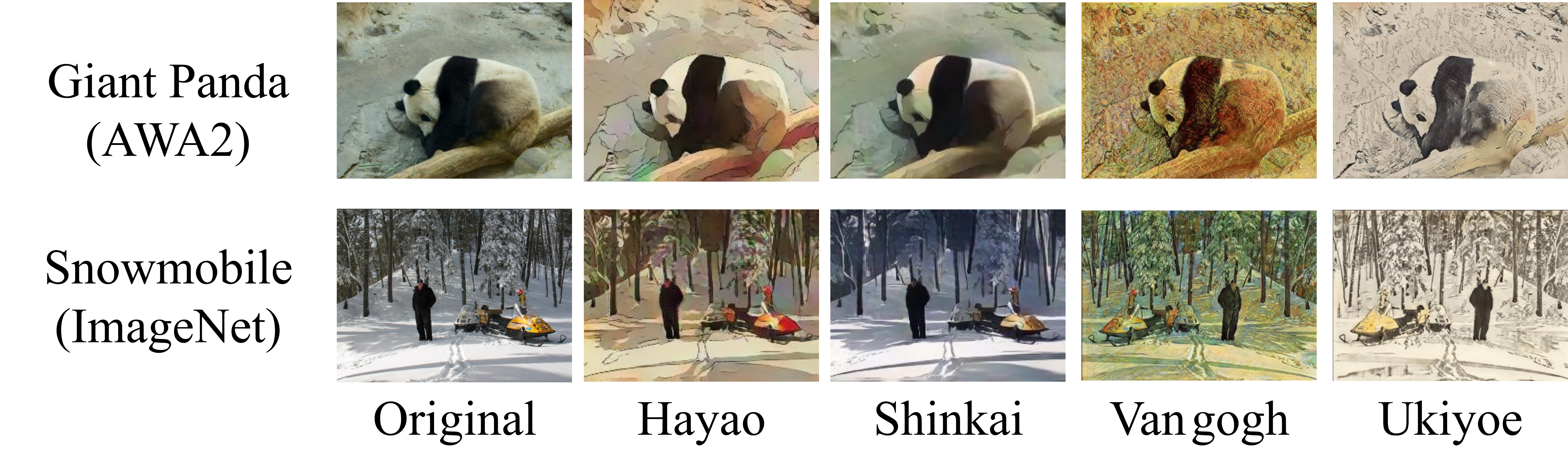}
\caption{\small Examples of simulated domain shifts by off-the-shelf style transfer models. Each original sample was randomly assigned a style to make up for the final dataset. Some samples of non-head classes in certain domains were randomly dropped out to simulate the practical \ltds{} problem studied in this work.}
\label{fig:dataset}
\end{minipage}
\hfill
\begin{minipage}[b!]{0.48\linewidth}
\centering
\includegraphics[width=0.9\linewidth]{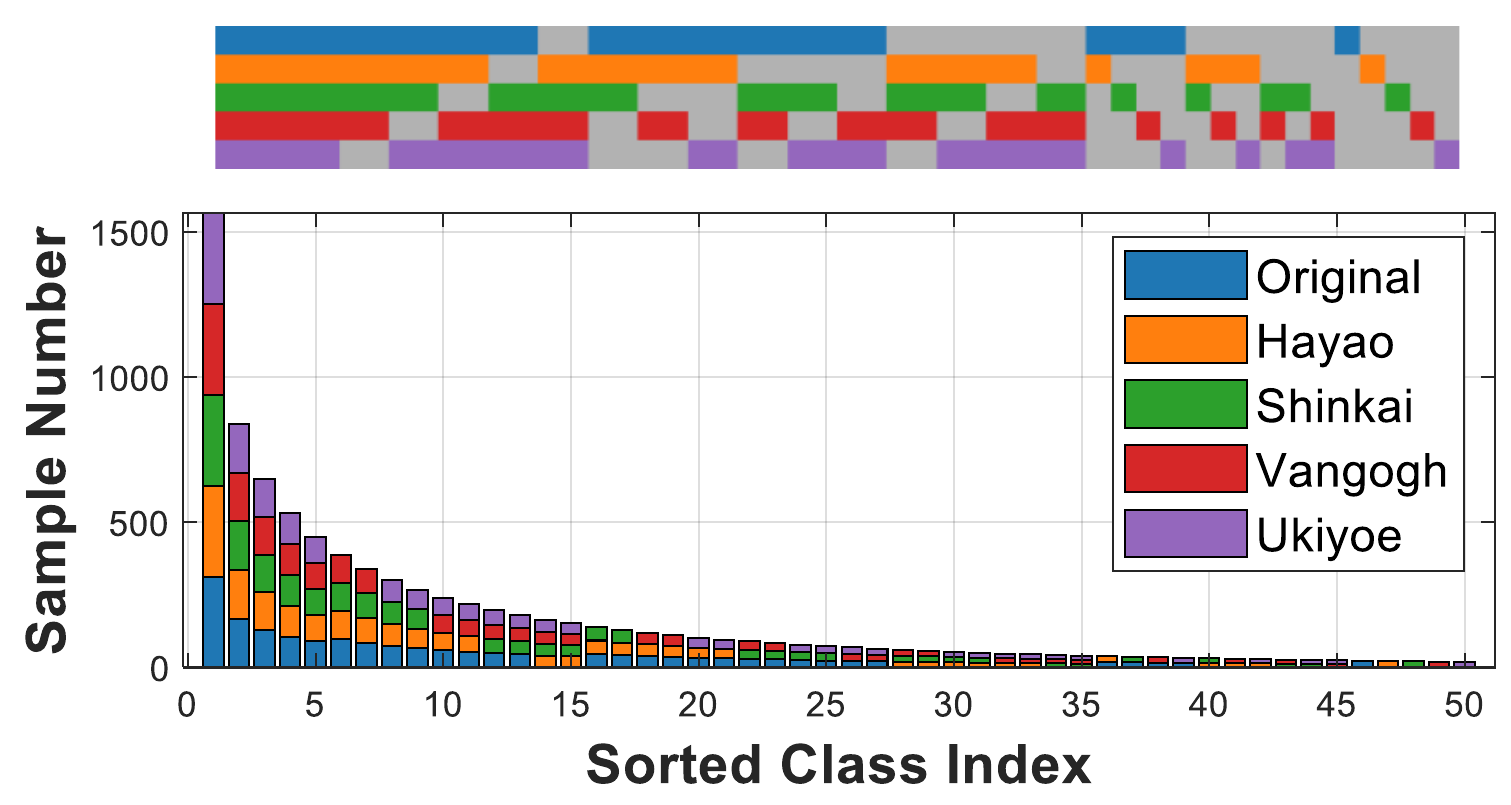}
\caption{\small Data distribution of \awalts{}.}
\label{fig:awa2}
\tabcaption{Dataset Details.}
\label{tab:dataset}
\setlength{\tabcolsep}{3pt}
\resizebox{0.9\columnwidth}{!}{
\small
\begin{tabular}{@{}l|cccc@{}}
    \toprule
    \textbf{Dataset} & \textbf{$|D|$} & \textbf{$|C|$} & \textbf{$|\mathbf{x}|$} & \textbf{Ratio*} \\
     \midrule
     \textbf{AWA2-LTS}  & 5 & 50 & $\sim$8k & $\sim$78\\
     \textbf{ImageNet-LTS} & 5 & 1000 & $\sim$100k & $\sim$256\\
     \bottomrule
\end{tabular}}
\end{minipage}
\hfill
\end{figure}

Two datasets were adopted to evaluate the effectiveness of our proposed methods. \awa{}~\cite{xian2018zero} and \imagenetlt{}~\cite{liu2019large}.  To benchmark our proposed task, \awa{} was firstly randomly resampled to convert to a long-tailed version. Regarding \textbf{AWA2-LT} and \textbf{ImageNet-LT}, we applied off-the-shelf style transfer models (Hayao~$\&$~Shinkai\footnotemark[2]; Vangogh~$\&$~Ukiyoe\protect\footnotemark[3]) to simulate domain shifts as shown in~\figurename~\ref{fig:dataset} and the generated new datasets are referred to as \textbf{AWA2-LTS} and \textbf{ImageNet-LTS}, respectively. Some classes from certain domains were deliberately dropped out to simulate a more realistic settings with entangled long-tailed and shifted distributions. A visual explanation of \awalts{} training data distribution is given in \figurename~\ref{fig:awa2}. 
Please see Table \ref{tab:dataset} and refer to supplementary material for more details.   
\footnotetext[2]{\url{https://github.com/Yijunmaverick/CartoonGAN-Test-Pytorch-Torch}}
\footnotetext[3]{\url{https://github.com/junyanz/pytorch-CycleGAN-and-pix2pix}}

\subsubsection{Model Training Setup.} We deployed all models using Pytorch with NVIDIA RTX3090. For \awalts{} and \imagenetlts{}, ResNet10 was adopted as the backbone with random initialization. Regarding semantic features, Word2Vec embeddings \cite{WORD2VEC} were adopted for \imagenetlt{}. BERT embeddings \cite{BERT} of class descriptors were utilized for \awalts{}. More details can be found in supplementary material and our project page \url{https://xiaogu.site/LTDS}.

\subsubsection{Evaluation Setup.}
Following \cite{shu2021open,you2019universal}, a threshold was empirically set up to apply on the prediction confidence. Classes below the given threshold are defined as open classes, belonging to $\mathcal{Y}\setminus\mathcal{Y}^{tr}$. We applied leave-one-domain-out protocol for evaluation. We reported the \textbf{\textit{Acc-U}} for the non-open classes in the held-out unseen domain. In addition, the domain-averaged accuracy of non-open classes in all the domains \textbf{\textit{Acc}}, and harmonic score \textbf{\textit{H}} of all the classes in all the domains are reported. The score \textbf{\textit{Acc}} and \textbf{\textit{H}} take into account those classes belonging to $\mathcal{Y}^{tr}\setminus\mathcal{Y}^{i}$ of each individual domain $i$. Therefore both metrics can to some extent reflect whether those domain-specific tail classes could lead to spurious correlations during training. 

\subsection{Long-Tail with Conditional Distribution Shifts}
We evaluated the performance under \ltds{} based on two benchmarks \awalts{} and \imagenetlts{} proposed in this work. For comparison, Agg Baseline, \lt{} solutions (cRT~\cite{kang2019decoupling}, BSCE~\cite{ren2020balanced}, Equal~\cite{tan2020equalization}, Remix~\cite{chou2020remix}), \textbf{DG} solutions (Epic-FCR~\cite{li2019episodic}, MixStyle~\cite{zhou2021mixstyle}, CuMix~\cite{mancini2020towards}, DAML~\cite{shu2021open}) were implemented.

\subsubsection{Quantitative Results.}
\begin{table}[t]
\centering
\tabcaption{Results on \awalts{} based on leave-one-domain-out evaluation.}
\label{tab:awa2}
\setlength{\tabcolsep}{3pt}
\resizebox{\columnwidth}{!}{
\begin{tabular}{llccccccccccccccc}
\toprule
 & &  \multicolumn{3}{c}{\textbf{Original}} & \multicolumn{3}{c}{\textbf{Hayao}} & \multicolumn{3}{c}{\textbf{Shinkai}}  & \multicolumn{3}{c}{\textbf{Vangogh}}   & \multicolumn{3}{c}{\textbf{Ukiyoe}} \\ \cmidrule(lr){3-5} \cmidrule(lr){6-8} \cmidrule(lr){9-11} \cmidrule(lr){12-14} \cmidrule(lr){15-17} 
 \multicolumn{2}{l}{\textbf{Method}} & Acc-U & Acc & H & Acc-U & Acc & H & Acc-U & Acc & H & Acc-U & Acc & H & Acc-U & Acc & H \\ \midrule
 & Agg  & 29.4 & 27.0 & 34.5 & 20.4 & 29.9 & 36.2 & 30.8 & 33.5 & 38.6 & 27.1 & 34.2 & 41.0 & 25.5 & 34.2 & 38.6 \\ \midrule
\multirow{4}{*}{\rotatebox{90}{\textbf{LT}}} 
  & cRT\cite{kang2019decoupling} & 30.4 & 29.1 & 34.8 & 23.5 & 33.6 & 38.9 & \best{34.7} & 35.8 & 39.6 & 28.6 & 35.8 & 43.0 & 28.4 & 36.7 & 35.7 \\
  & BSCE\cite{ren2020balanced} & 41.8 & 35.9 & 41.7 & 24.7 & 36.1 & 41.7 &  30.2 & 35.8 & 40.7 & 29.0 & 37.7 & 43.2 & 25.9 & 33.6 & 35.1  \\
  & Equal\cite{tan2020equalization} & 34.1 & 32.9 & 36.6 & 24.3 & 35.3 & 42.7 & 33.5 & 36.2 & 40.5 & 28.8 & 35.8 & 42.4 & 27.3 & 34.7 & 34.0 \\ 
  & Remix\cite{chou2020remix} & 32.7 & 30.3 & 35.9 & 16.9 & 30.7 & 33.3 & 27.6 & 32.0 & 37.5 & 26.9 & 31.8 & 41.2 & 26.5 & 32.0 & 34.9 \\ \midrule
\multirow{6}{*}{\rotatebox{90}{\textbf{DG}}} & Epi-FCR\cite{li2019episodic} & 34.0 & 33.1 & 40.5 & 23.3 & 34.0 & 40.6 & 29.7 & 35.5 & 39.1 & 27.5 & 36.1 & 42.0 & 27.0 & 35.7 & 38.0  \\
   & MixStyle\cite{zhou2021mixstyle} &36.7 & 34.0 & 41.2 & 27.1 & 36.2 & 41.7 & 32.0 & 36.2 & 40.6 & 28.4 & 36.0 & 41.9 & 28.8 & 36.2 & 38.3 \\
   & CuMix\cite{mancini2020towards} & 36.1 & 33.8 & 38.6 & 24.7 & 35.3 & 41.0 & 30.2 & 35.1 & 41.4 & 28.2 & 35.1 & 40.9 & 26.5 & 34.7 & 34.6 \\ 
   & DAML\cite{shu2021open} & 13.9 & 10.7 & 16.2 &  14.7 & 22.5 & 29.8 & 17.3 & 24.9 & 30.4 & 14.3 & 19.5 & 25.6 & 22.9 & 28.9 & 36.0 \\ 
    & DAML\cite{shu2021open}-Warmup & \sec{42.2} & 35.3 & \sec{42.5} & 25.7 & 35.2 & 39.5 & 31.2 & 36.8 & 44.0 & 29.4 & 37.5 & \sec{45.3} & 28.6 & 36.0 & \best{41.4} \\ \midrule
 & MixStyle+BSCE & 40.0 & 36.8 & 41.8 & \sec{28.8} & \sec{39.7} & \sec{43.5} & 32.4 & 38.3 & \sec{44.2} & \sec{30.8} & 38.2 & 43.3 & \sec{29.8} & \sec{38.9} & \sec{39.3} \\ 
 & Epi-FCR+BSCE & 41.3 & \sec{36.9} & 42.0 & 24.0 & 35.9 & 41.2 & 32.0 & \sec{39.2} & 42.5 & 30.1 & \sec{38.5} & 41.7 & 26.6 & 35.9 & 38.7 \\
  \midrule
  & Ours  & \best{49.4} & \best{42.1} & \best{45.8} & \best{29.8} & \best{42.4} & \best{46.3} & \sec{34.3} & \best{42.6} & \best{45.3} & \best{32.7} & \best{40.3} & \best{46.3} & \best{32.9} & \best{42.4} & \sec{39.3} \\
\bottomrule
\end{tabular}}
\end{table}
As shown in \tablename s~\ref{tab:awa2} \& \ref{tab:imagenet}, our method achieves overall superior performance compared to other methods. In addition, two combinations MixStyle+BSCE and Epi-FCR+BSCE were applied on~\awalts{} dataset for comparison. We noticed the extremely low performance of DAML~\cite{shu2021open} under \ltds{}. Since it is based on the knowledge distillation from each domain-specific model, the knowledge learned from individual long-tailed distribution would be significantly biased towards the head classes. We simply moderated such bias by a warm-up pretraining, and the variant is referred to as DAML-Warmup. Although DAML-Warmup can alleviate the class imbalance issue when the imbalance ratio and class number is small, we observed its failure on \imagenetlts{}. It may indicate that it cannot handle the semantic information when a large proportion of classes are missing.
Overall, our results demonstrate superior performance compared to existing methods.

\begin{table}[t]
\centering
\tabcaption{Results on \imagenetlts{} based on leave-one-domain-out evaluation.} 
\label{tab:imagenet}
\setlength{\tabcolsep}{3pt}
\resizebox{\columnwidth}{!}{
\begin{tabular}{llccccccccccccccc}
\toprule
 & &  \multicolumn{3}{c}{\textbf{Original}} & \multicolumn{3}{c}{\textbf{Hayao}} & \multicolumn{3}{c}{\textbf{Shinkai}}  & \multicolumn{3}{c}{\textbf{Vangogh}}   & \multicolumn{3}{c}{\textbf{Ukiyoe}} \\ \cmidrule(lr){3-5} \cmidrule(lr){6-8} \cmidrule(lr){9-11} \cmidrule(lr){12-14} \cmidrule(lr){15-17} 
 \multicolumn{2}{l}{\textbf{Method}} & Acc-U & Acc & H & Acc-U & Acc & H & Acc-U & Acc & H & Acc-U & Acc & H & Acc-U & Acc & H \\ \midrule
 & Agg & 19.5 & 18.1 & 23.5 & 13.0 & 18.2 & 24.4 & 15.2 & 18.0 & 24.2 & 13.2 & 17.5 & 23.7 & 12.5 & 17.9 & 24.1  \\  \midrule
\multirow{4}{*}{\rotatebox{90}{\textbf{LT}}} 
  & cRT\cite{kang2019decoupling} & 20.7 & 18.7 & 24.8 & 13.8 & 19.0 & 24.3 & 16.2 & 18.8 & 24.8 & 14.0 & 18.4 & \sec{25.5} & 13.6 & 18.7 & 25.3 \\
  & BSCE\cite{ren2020balanced} & \sec{20.8} & \sec{19.1} & \sec{25.6} & \sec{14.3} & \sec{19.4} & \sec{25.2} & \sec{16.5} & 19.1 & \sec{25.2} & \best{14.7} & \sec{18.8} & \sec{25.5} & \sec{14.1} & \sec{19.1} & \sec{26.4}\\
  & Equal\cite{tan2020equalization} & 16.3 & 15.4 & 20.6 & 10.7 & 15.2 & 20.6 & 13.2 & 16.4 & 22.2 & 10.5 & 14.8 & 21.4 & 10.9 & 15.8 & 21.3 \\ 
  & Remix\cite{chou2020remix} & 14.8 & 13.8 & 18.6 & 10.1 & 14.1 & 20.0 & 11.3 & 14.1 & 22.0 & 11.1 & 13.5 & 18.2 & 10.5 & 14.8 & 20.8 \\ \midrule
\multirow{4}{*}{\rotatebox{90}{\textbf{DG}}} & Epi-FCR\cite{li2019episodic} & 19.2 & 18.8 & 23.1 & 13.5 & 19.0 & 23.3 & 15.0 & \sec{20.0} & 25.1 & 12.2 & 18.0 & 23.8 & 13.0 & 17.5 & 25.2 \\
   & MixStyle\cite{zhou2021mixstyle} & 17.7 & 16.4 & 22.0 & 12.1 & 16.7 & 22.1 & 13.6 & 16.5 & 22.9 & 11.8 & 16.0 & 22.9 & 11.5 & 16.2 & 21.4 \\
   & CuMix\cite{mancini2020towards} & 18.2 & 17.2 & 23.7 & 13.2 & 17.5 & 22.7 & 14.2 & 17.1 & 22.4 & 12.1 & 16.8 & 22.3 & 12.1 & 17.1 & 23.7 \\
   & DAML\cite{shu2021open}-Warmup & 14.7 & 12.7 & 18.2 & 10.5 & 13.0 & 18.2 & 11.5 & 13.2 & 18.0 & 9.7 & 12.8 & 16.9 & 10.1 & 13.0 & 18.6 \\ \midrule
  & Ours  & \best{24.3} & \best{20.8} & \best{25.9} & \best{16.3} & \best{21.3} & \best{25.4} & \best{17.4} & \best{20.9} & \best{25.8} & \sec{14.3} & \best{20.3} & \best{26.6} & \best{15.4} & \best{20.3} & \best{26.7} \\
\bottomrule
\end{tabular}}
\end{table}

\subsubsection{Qualitative Results.}
We present the t-SNE visualizations of features from the test split of \awalts{} in \figurename~\ref{fig:tsne} along with the results from the baseline Agg. Different colors indicate different domains on the upper row, whereas different categories ranging from head to tail on the bottom. It can be observed that the samples which are sampled from the same classes but from different domains (including the unseen domain) are better clustered. More qualitative analysis can be found in supplementary material. 

\subsubsection{Ablation Studies.}
\begin{figure}[btp!]
\begin{minipage}[b!]{0.60\linewidth}
\centering
\tabcaption{Ablation studies on \awalts{}. CE indicates cross-entropy loss.}
\label{tab:awa2_ablation}
\setlength{\tabcolsep}{3pt}
\resizebox{\columnwidth}{!}{
\begin{tabular}{ccccccccccc}
\toprule     
     \multirow{2}{*}{\textbf{Index}} &\multirow{2}{*}{$\mathcal{L}_{dc}$} & \multirow{2}{*}{CE} & \multirow{2}{*}{$\mathcal{L}_{Z2S}$} & \multirow{2}{*}{$\mathcal{L}_{S2S}$} & \multirow{2}{*}{$\mathcal{L}_{S2Z}$} & \multirow{2}{*}{$\mathcal{L}_{Aug}$} & \multirow{2}{*}{Meta} & \multicolumn{3}{c}{\textbf{Avg}} \\ \cmidrule(lr){9-11}
    & & & & & & & & Acc-U & Acc & H \\\midrule
a &  -      & \cmark & -      & -      & -      & -      & -      & 26.6 & 31.8 & 37.8 \\
b &  \cmark & -      & -      & -      & -      & -      & -      & 30.6 & 36.7 & 39.2 \\
c &  -      & \cmark & -      & -      & -      & -      & \cmark & 28.9 & 34.3 & 33.8 \\
d &  \cmark & -      & -      & -      & -      & -      & \cmark & 34.3 & 40.0 & 40.2 \\
e &  \cmark & -      & \cmark & -      & -      & -      & -      & 31.5 & \sec{41.2} & 41.0 \\
f &  \cmark & -      & \cmark & \cmark & -      & -      & -      & 32.0 & 40.8 & 41.2 \\
g &  \cmark & -      & \cmark & \cmark & \cmark & -      & -      & 32.5 & 40.8 & {42.8} \\
h &  \cmark & -      & -      & -      & -      & \cmark & -      & 26.2 & 30.0 & 36.8 \\
i &  \cmark & -      & \cmark & \cmark & \cmark & \cmark & -      & 34.0 & 40.9 & {42.8} \\
j &  \cmark & -      & \cmark & \cmark & \cmark & \cmark & \cmark & \best{35.7} & \best{42.0} & \best{44.6}  \\ \midrule
k & \multicolumn{7}{l}{Single Prototype Alignment} & 32.5 & 41.0 & 41.6 \\
l & \multicolumn{7}{l}{No-Weight Augmentation} & \sec{34.6} & 40.7 & \sec{43.0} \\ \bottomrule
\end{tabular}}
\end{minipage}
\hfill
\begin{minipage}[b!]{0.38\linewidth}
\centering
\includegraphics[width=0.45\linewidth, height=0.3\linewidth]{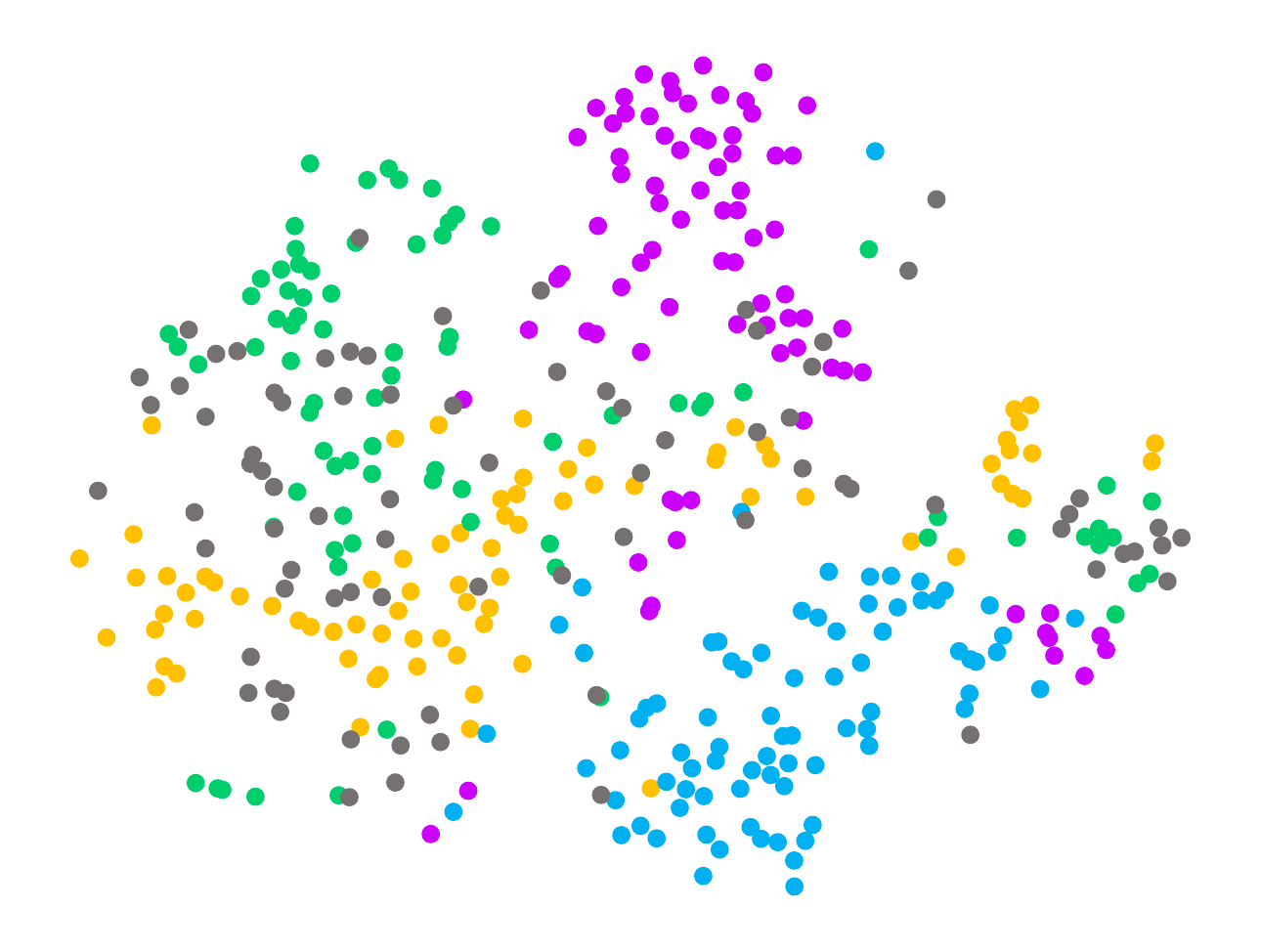}
\includegraphics[width=0.45\linewidth, height=0.3\linewidth]{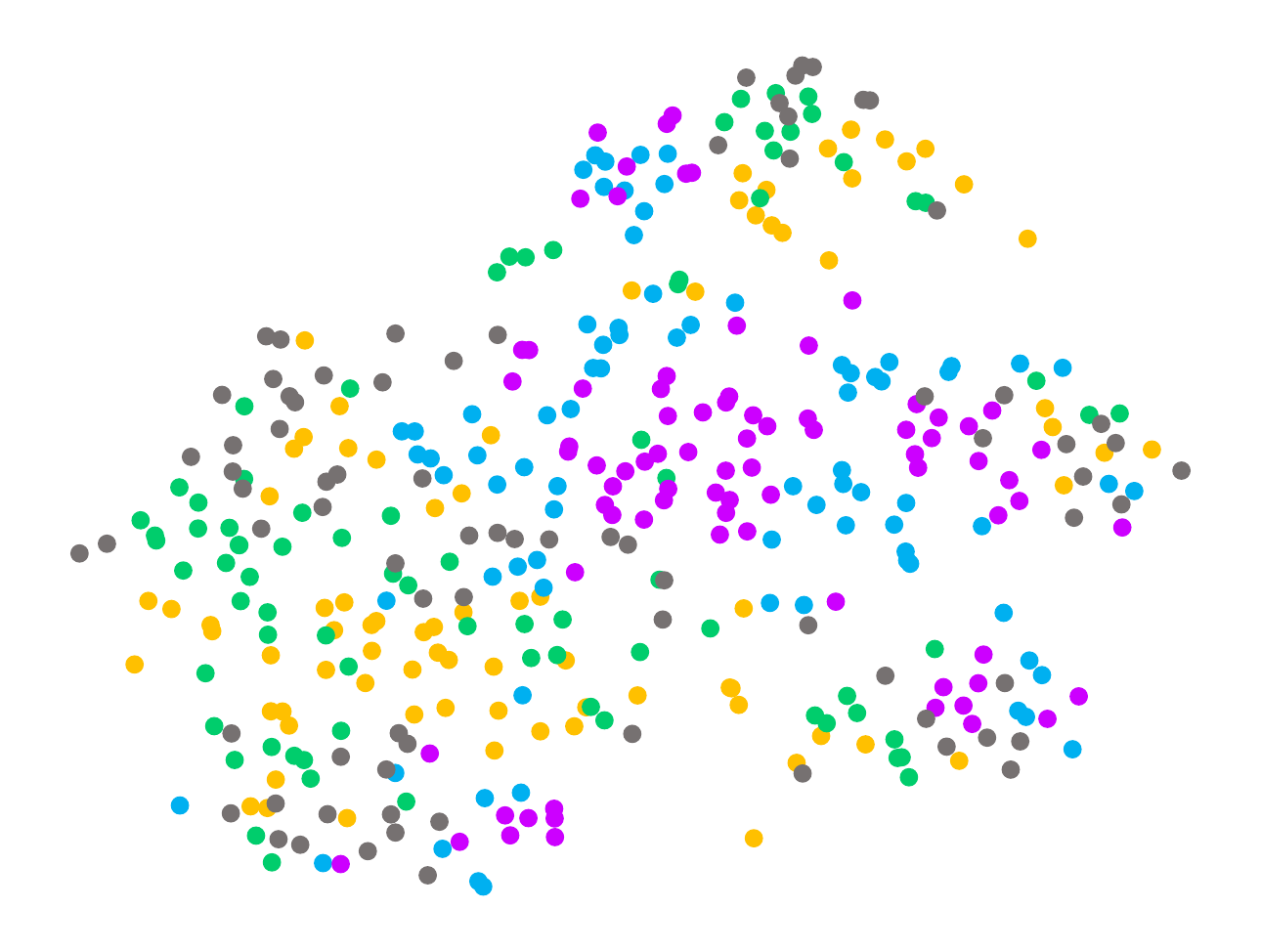}
\includegraphics[width=0.5\linewidth]{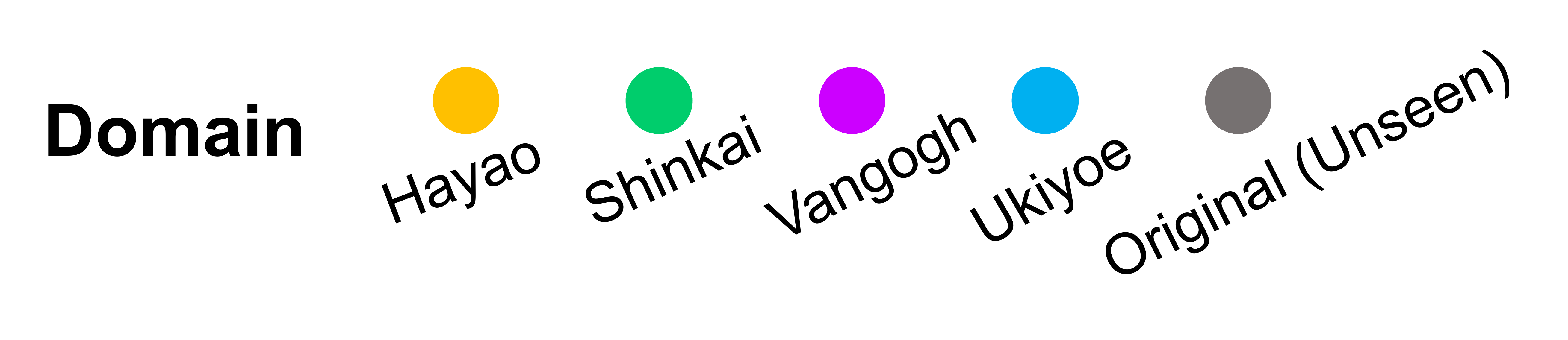} \\
\includegraphics[width=0.45\linewidth, height=0.3\linewidth]{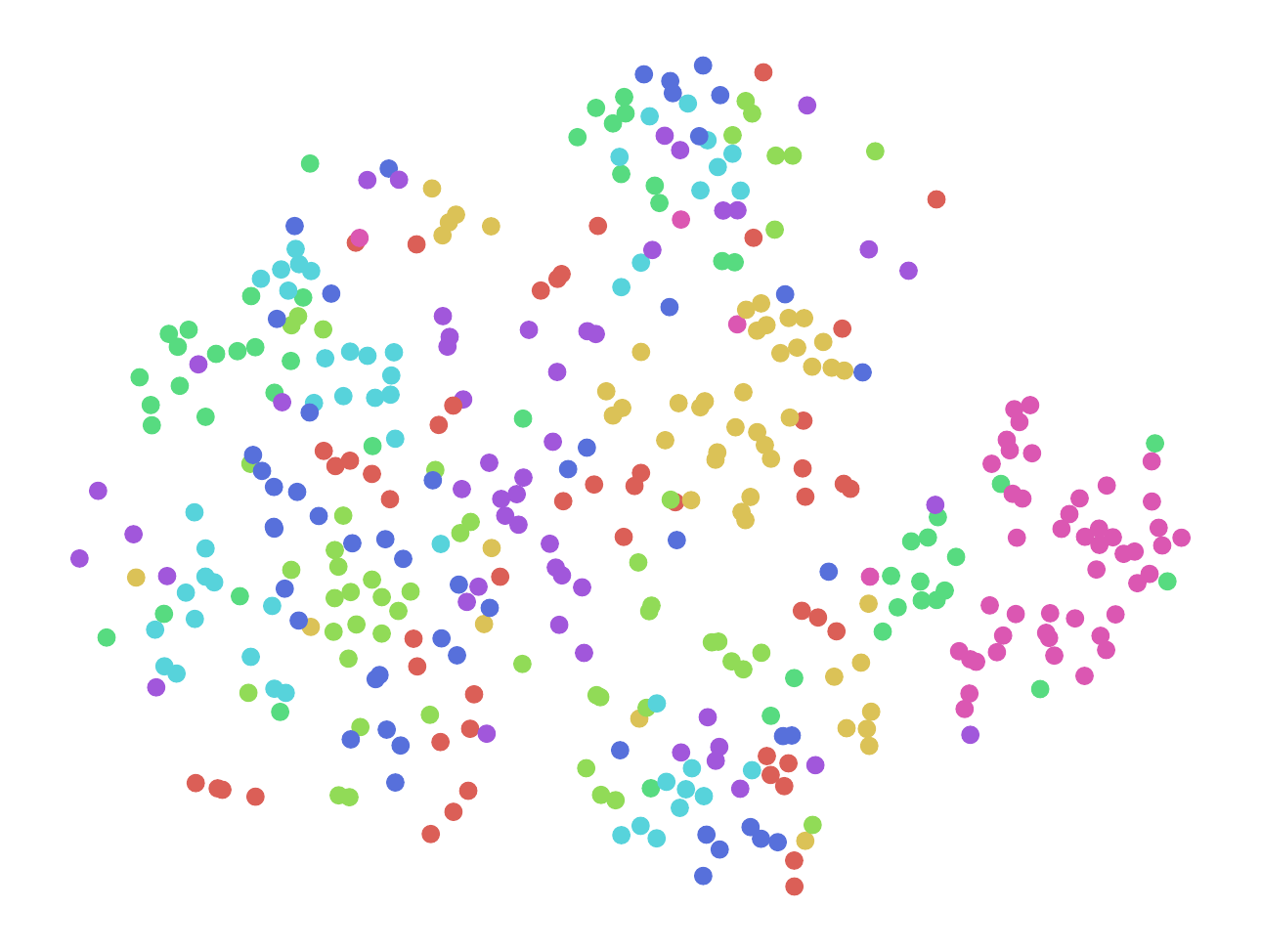}
\includegraphics[width=0.45\linewidth, height=0.3\linewidth]{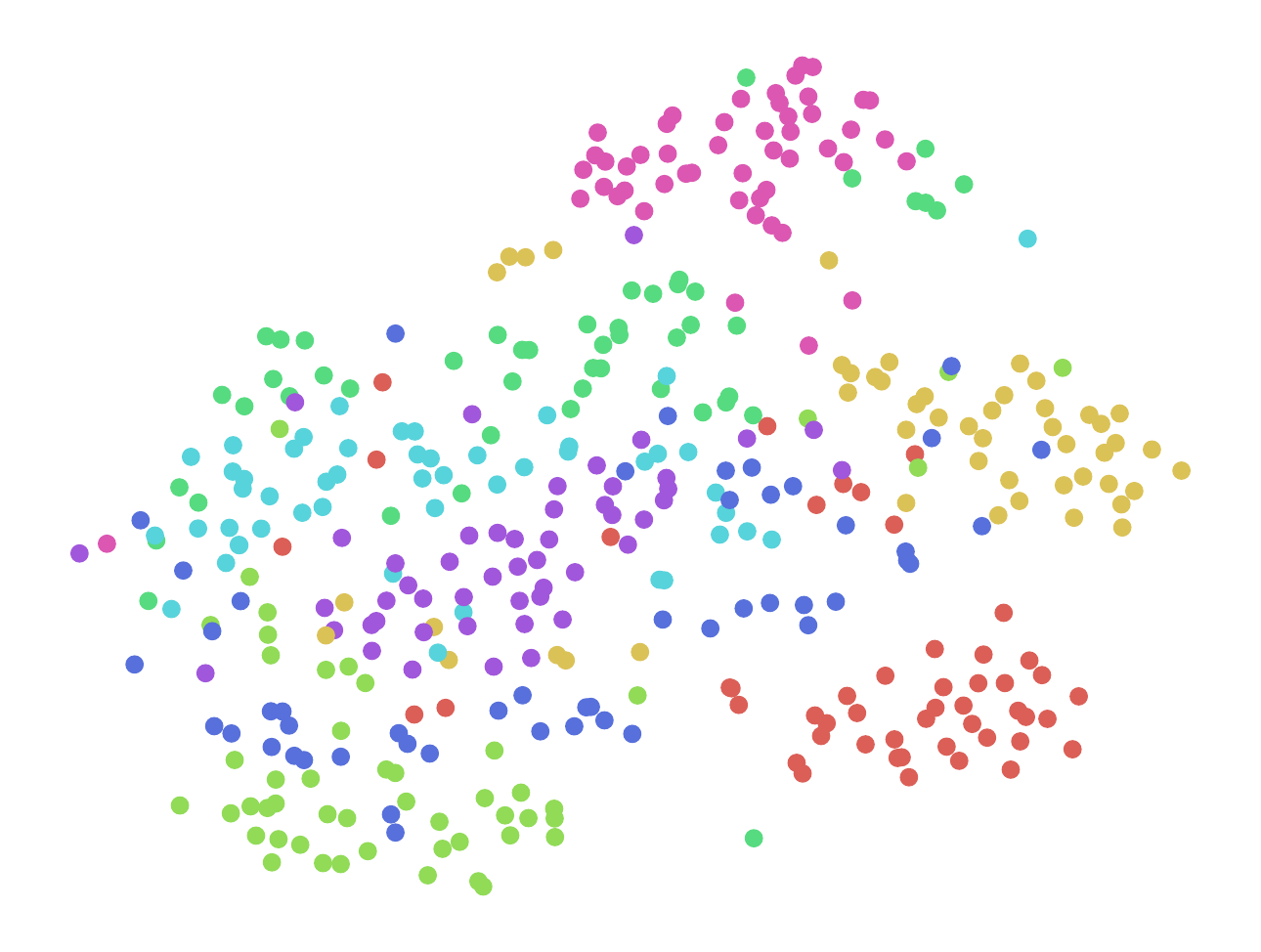}
\includegraphics[width=0.5\linewidth]{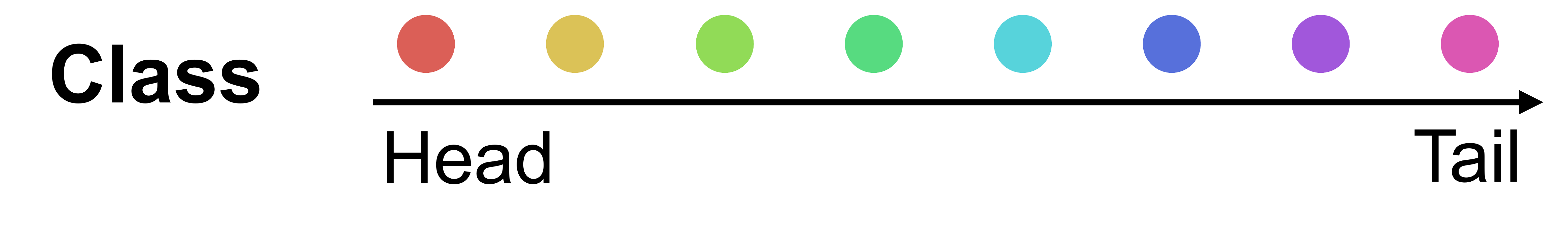}\\
{\tiny (a) Agg \qquad \qquad \qquad (b) Ours}
\figcaption{t-SNE of \awalts{} test set. Left: Agg. Right: Ours.}
\label{fig:tsne}
\end{minipage}
\hfill
\end{figure}

We validate the performance of each proposed module and the whole meta learning framework by ablation studies as shown in \tablename~\ref{tab:awa2_ablation} indicates the effectiveness of each individual modules. In particular, the ablated model indexed by d also presents relatively good performance, demonstrating that even without additional semantic features for alignment, it is still comparable to existing solutions. We also did two another ablation studies, with just single prototype for alignment and without weighted term in \equationautorefname~\eqref{eq:aug} for updating covariance matrix. Please see more detailed discussions in our supplementary material. Additional experiments on sample complexity, choices of embeddings, as well as results on open domain generalization datasets can also be found in supplementary materials.   

\section{Conclusions}
Long-tailed category distribution and domain shifts have been two major issues concerned with real-world datasets, leading to degraded performance upon practical deployment. The combination of these two problems poses a significantly challenging scenario, where not only these two problems should be addressed, but the domain-specific non-head classes should also be paid attention to, to avoid shortcuts. We proposed a meta-learning framework to ensure that the model can perform well over all classes and all domains, including unseen novel domains. We evaluated two benchmarks proposed in this paper. The experimental results demonstrate that the proposed method can achieve superior performance, when compared to either long-tailed/domain-generalization solutions or the combinations. In the future, we are going to apply our method to more specific applications like behavioural analysis and health care.

\par\vfill\par

\clearpage
%
%
\bibliographystyle{splncs04}

\clearpage
\section*{Supplementary Materials}
\setcounter{table}{0}
\renewcommand{\thetable}{S\arabic{table}}
\setcounter{figure}{0}
\renewcommand{\thefigure}{S\arabic{figure}}
\setcounter{section}{0}
\renewcommand\thesection{\Alph{section}}
\setcounter{equation}{0}
\renewcommand{\theequation}{S\arabic{equation}}

\section{Experiment Settings}
\subsection{Datasets}
\subsubsection{\awalts{} and \imagenetlts{}:} These two datasets were modified from the well-established \awa{}~\cite{xian2018zero} and \imagenetlt{}~\cite{liu2019large} datasets. The former (50 categories) is a benchmark for zero-shot learning, whilst the latter (1000 categories) has been widely used for long-tailed visual recognition. To accommodate our practical needs for benchmarking \ltds{}, the following steps were applied. 

First of all, for~\imagenetlt{}, the training/validation/testing splits are defined in \cite{liu2019large}, where the training subset is long-tailed distributed whilst validation and testing ones are balanced sampled. Our further processing of each subset was based on the original splits. For \awa{}, we randomly extracted 50 samples of each category for testing, and 30 samples for validation. Since the training data of \awa{} is not long-tailed originally, we performed random sampling to convert the training subset of \awa{} to a long-tailed version with the maximum number 1565, and the minimum number 20, following $n_c=\left \lfloor  n_{max}\exp(-\frac{\sqrt{c-1}}{7} \times \log\frac{n_{min}}{n_{max}}) \right \rfloor$, where $c$ refers the sorted class index, starting from 1 to the class number $C=50$. 

\begin{figure}[h!]
    \centering
    \subfigure[\awalts{}\label{fig:awa2-sm}]{\includegraphics[width=0.45\linewidth]{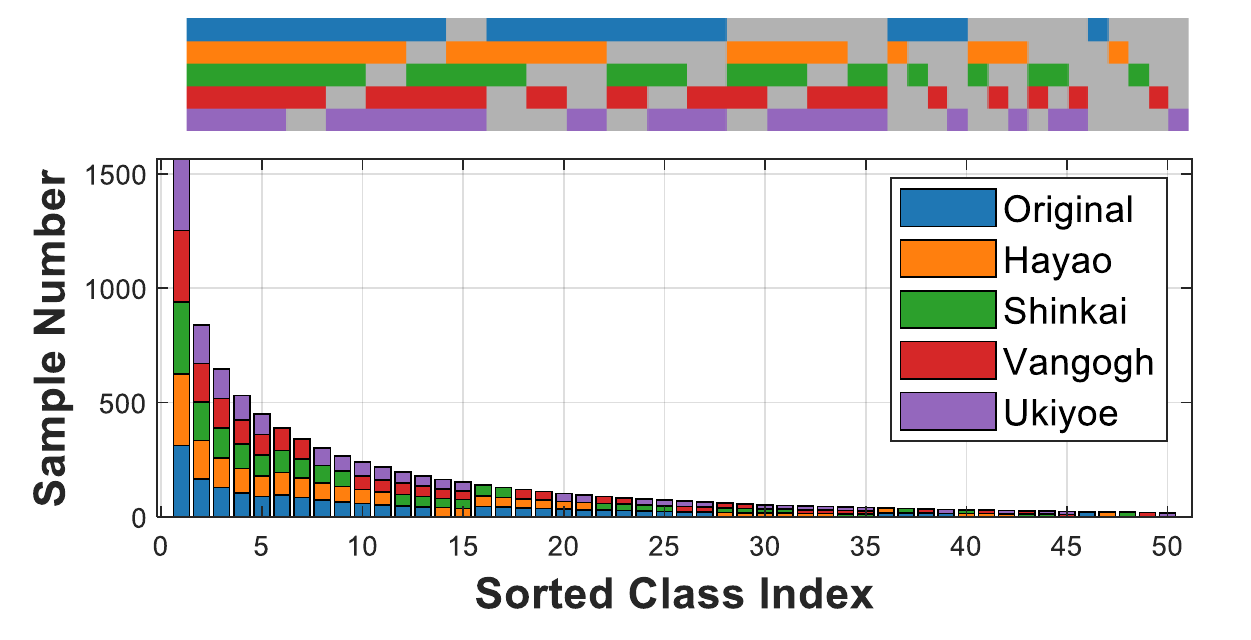}}
    \subfigure[\imagenetlts{}\label{fig:imagenet-sm}]{\includegraphics[width=0.45\linewidth]{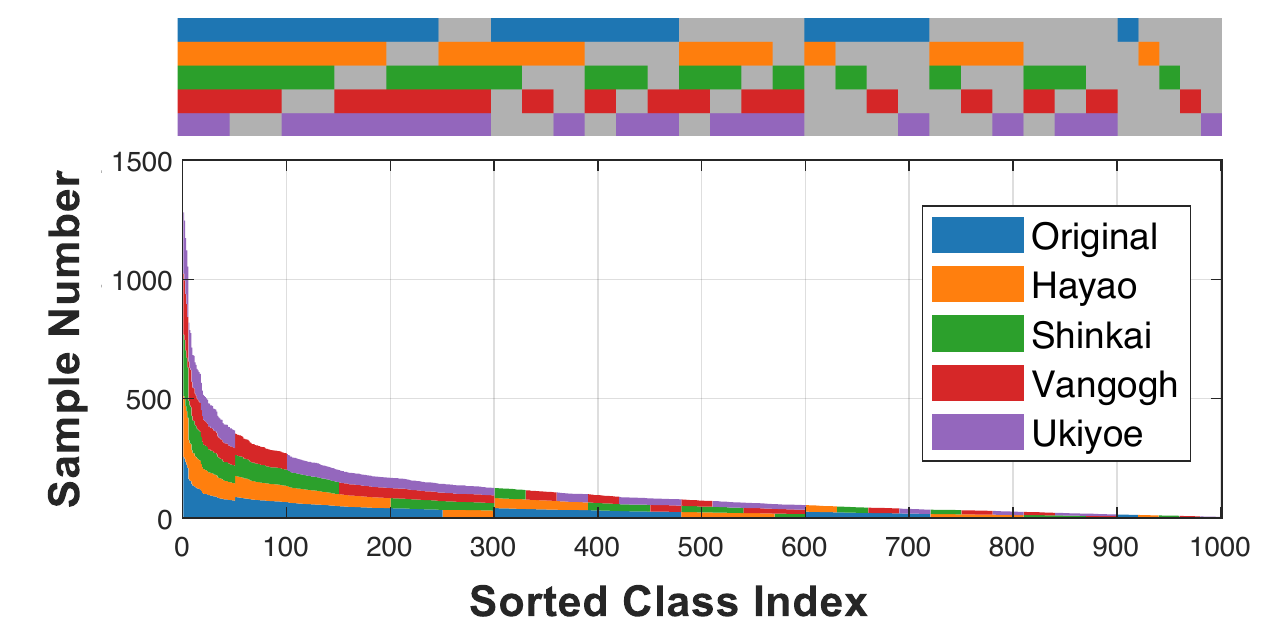}}    
    \caption{Categorical Distribution of Training Subset of Each Domain. Left: \awalts{}; Right: \imagenetlts{}. The bottom row shows the categorical distribution across domains and categories, whereas the top row presents the existence of classes in the training subset of each domain (grey indicates non-existence).}
\end{figure}

Subsequently, for the training data, each image was assigned a style randomly selected from the five ones (i.e., Original, Hayao, Shinkai, Vangogh, Ukiyoe). To simulate the realistic scenario where head classes are common across domains whilst non-head classes appear in only certain specific domains due to their low-frequency, we deliberately reduced the number of domain candidates for those non-head classes. The categorical distributions of the training subset in each domain are shown in Figs.~\ref{fig:awa2-sm} and~\ref{fig:imagenet-sm}. 

Regarding the validation subset, after assigning each image with a style with equal probability, we only kept those seen classes in the training subset of each individual domain. This ensures that the training label set and validation label set of each domain are totally the same. Regarding the testing subset, each image was assigned with a style randomly picked from five styles, and all classes were kept, so that all the classes are seen in each domain.

In total, for \awalts{}, totally 50 classes and 5 domains exist, whereas 1000 classes and 5 domains for \imagenetlts{}. Detailed instructions for generating the proposed \ltds{} datasets and the indexes of corresponding training/validation/testing splits for all images for benchmarking can be found at \url{https://github.com/guxiao0822/LT-DS/tree/main/dataset}. 

\subsection{Implementation Details}
\subsubsection{Architectures:}
For \awalts{} and \imagenetlts{}, we applied the ResNet-10 as the feature extractor $f$, and a fully connected (FC) layer as the classifier $h$. For $d$ and $e$, they are composed of a FC , BatchNorm, and ReLU layer. The whole network was randomly initialized without applying pretrained weights, so as to avoid the overlap of classes between our proposed datasets and the original ImageNet. %

\subsubsection{Training Details:}
We utilized SGD for optimization. The random seed was set as 0 for reproduction purposes. The parameters used in this paper was listed as in \tablename~\ref{tab:hp}. In addition, $\beta_2$ were decayed by 0.1 after 40 and 80 epochs.
\begin{table}[]
    \centering
    \caption{Hyperparameters (HP) in our experimental settings.}
    \label{tab:hp}
    \setlength{\tabcolsep}{3pt}
\resizebox{\columnwidth}{!}{
    \begin{tabular}{ll|ccc}
    \toprule
     \textbf{HP} & \textbf{Description} & \awalts{} & \imagenetlts{}  \\
     \midrule
     $\beta_1$ & meta-train learning rate & 0.2 & 0.2  \\
     $\beta_2$ & final learning rate &  0.1 & 0.1  \\
     k & top k similarity & 5 & 5 & - \\
     $\lambda$ & augmentation intensity &  5 & 5  \\
     B &  batch size of each domain & 48 & 64  \\
     $\alpha$ & margin of contrastive loss & 0.1 & 0.1  \\
     $\tau$ & temperature scaling constant & 1/30 & 1/50  \\
     $T_{max}$  & maximum training step (corresponding epoch number) & 100 & 100  \\
     $T_{\Sigma}$  & covariance tracking milestone step (corresponding epoch number) &  40 & 40 \\
     $w_1$ & weight of $\mathcal{L}_{Z2S}$  & 0.1 & 0.1 \\
     $w_2$ & weight of $\mathcal{L}_{S2S}$  & 0.1 & 0.1 \\
     $w_3$ & weight of $\mathcal{L}_{S2Z}$  & 0.1 & 0.1 \\
     $w_4$ & weight of $\mathcal{L}_{Aug}$ and $\mathcal{L}_{MAug}$& 0.1 & 0.1 \\
     $w_{mte}$ & weight of $\mathcal{L}_{mte}$ & 0.3 & 0.3  \\
    \bottomrule
    \end{tabular}}
    \vspace{-10pt}
\end{table}

For the compared methods, we used the same backbone for fair comparison, with the hyperparameter settings adopted in their original implementations.



\section{Equation Proof}
\subsection{Distribution Calibrated Classification Loss}

The distribution calibrated classification loss aims to calibrate the classification loss to a balanced category distribution, so that meta-train and meta-test both aim to achieve ideal performance on balanced distributions.  
\begin{equation}
   \small
    \mathcal{L}_{dc}(\boldsymbol{x}_i, y_i, d_i; f, h) = -\log\frac{n_{y_i}^{d_i} \exp{([h\circ f(\boldsymbol{x}_i)]_{y_i}})}{\sum_{c=1}^C n_c^{d_i} \exp({[h\circ f(\boldsymbol{x}_i)]_{c})}}.
   \label{eq:dc-sm}
\end{equation}

Without loss of generality and for simplicity, we denote the logits of class $i$ as $\gamma_i=[h\circ f(\boldsymbol{x})]_{i}$. The probability of class $\phi_i$ after softmax normalization is $\frac{\exp(\gamma_i)}{\sum_{c} \exp(\gamma_c)}$.

Based on Bayesian theorem, the probability $\phi^d_i$ in domain $d$ is formulated as below, 

\begin{equation}
\small
    \phi^d_i = p^d(y=i|f(\boldsymbol{x})) = \frac{p^d(f(\boldsymbol{x})|y=i)p^d(y=i)}{p^d(f(\boldsymbol{x}))}.
\end{equation}

For another domain $d^\prime$, 
\begin{equation}
    \phi^{d^\prime}_i =\frac{p^{d^\prime}(f(\boldsymbol{x})|y=i)p^{d^\prime}(y=i)}{p^{d^\prime}(f(\boldsymbol{x}))}.
\end{equation}

Considering our visual-semantic mapping functional blocks, we assume that the term $\frac{p^d(f(\boldsymbol{x})|y=i)}{p^d(f(\boldsymbol{x}))}$ is identical across domains. Therefore, 

\begin{align}
\small
    \phi_i^{d^\prime} &= {\phi_i^d}\frac{p^{d^\prime}(y=i)}{p^{d}(y=i)}, \\
    &= \frac{\exp(\gamma_i)}{\sum_{c} \exp(\gamma_c)} \frac{n^{d^\prime}_i/\sum_c{n^{d^\prime}_c}}{n^d_i/\sum_c{n^{d}_c}},\\
    &= \frac{\sum_c{n^{d}_c}}{\sum_c{n^{d^\prime}_c}\sum_{c} \exp(\gamma_c)}\frac{n^{d^\prime}_i\exp(\gamma_i)}{n^d_i}. \label{eq:7}
\end{align}

Based on the property that the summed probability over all classes equals to one, we can derive

\begin{align}
\small
\sum_{i=1}^C \phi_i^{d^\prime} =  \frac{\sum_c{n^{d}_c}}{\sum_c{n^{d^\prime}_c}\sum_{c}\exp(\gamma_c)} \sum_{i=1}^C \frac{n^{d^\prime}_i\exp(\gamma_i)}{n^d_i} = 1. \label{eq:8}
\end{align}

Then, based on Equations~\eqref{eq:7} and~\eqref{eq:8}, it can be derived that
\begin{align}
    \phi_i^{d^\prime} &= \frac{\frac{n^{d^\prime}_i}{n^d_i}\exp(\gamma_i)}{\sum_{c}\frac{n^{d^\prime}_c}{n^d_c}\exp(\gamma_c)}.
\end{align}

In our case, $d^\prime$ is recognized as a training domain with imbalanced distribution and $d$ as a testing domain with balanced distribution. Since $n_c^d$ are equal across classes, the probability of class $i$ in $d^\prime$ can be rewritten as below, 

\begin{align}
    \phi_i^{d^\prime} &= \frac{n^{d^\prime}_i\exp(\gamma_i)}{\sum_{c} n^{d^\prime}_c\exp(\gamma_c)}.
\end{align}

Till now, Equation~\eqref{eq:dc-sm} is sorted. It should be noted that we do not calibrate the classification loss from meta-train domain category distributions to meta-test distributions. This is because each individual domain exists unseen classes (impossible to be rebalanced), and however our final goal is to achieve good performance over all classes. Instead, for both meta-train and meta-test, we aim to calibrate the classification loss to a balanced distribution. 

\subsection{Augmentation Loss}
Below are the surrogate loss for implicitly augmenting the feature diversity.

Denote the distribution of class $c$ to be a multivariate Gaussian distribution, where $f(\boldsymbol{x})$ of class $c$ obeys $\mathcal{N}(\boldsymbol{\mu}_c, \mathbf{\Sigma}_c)$. Consider a feature $f(\boldsymbol{x}_i)$ sampled along a direction from  $\mathcal{N}(\boldsymbol{\mu}_c, \lambda \mathbf{\Sigma}_c)$, where $\lambda$ indicates the augmentation intensity. Then the upper bound of classification loss can be viewed as below, 

\begin{align}
 \scriptsize
    & \mathbb{E}_{f(\boldsymbol{x}_i)} \Big [-\log\frac{\exp(w_{y_i}^\intercal f(\boldsymbol{x}_i) + b_{y_i})}{\sum_{c=1}^C \exp(w_c^\intercal f(\boldsymbol{x}_i) + b_c)} \Big ] \\
     &= \mathbb{E}_{f(\boldsymbol{x}_i)}\Big [\log\sum_{c=1}^C \exp((w_c^\intercal-w_{y_i}^\intercal)f(\boldsymbol{x}_i) + (b_c-b_{y_i})) \Big ] \\
     & \leqslant \log\Big [\sum_{c=1}^C \mathbb{E}_{f(\boldsymbol{x}_i)} \exp((w_c^\intercal-w_{y_i}^\intercal){f(\boldsymbol{x}_i)} + (b_c-b_{y_i})\Big ] \label{eq:ineq}\\
     & = \log \Big [\sum_{c=1}^C \exp((w_c^\intercal-w_{y_i}^\intercal)\mathbf{\boldsymbol{\mu}}_{y_i} + (b_c-b_{y_i}) + \frac{\lambda}{2}(w_c^\intercal-w_{y_i}^\intercal) \mathbf{\Sigma}_{y_i} (w_c - w_{y_i}))\Big ], \label{eq:est}
\end{align} 
where Equation~\eqref{eq:ineq} is derived based on the convex property of $\log$, whilst Equation~\eqref{eq:est} is based on the property that $\mathbb{E}[\exp(tX)]=\exp(t\mu+\frac{1}{2}\sigma^2t^2), X\sim \mathcal{N}(\mu, \sigma^2).$ It should be noted that the bias $[b_1, b_2, ..., b_C]^T$ was not considered in the main text for simplicity, yet was taken into account during our practical implementations. 

\subsubsection{Online Estimation of Visual Feature Prototype and Feature Covariances.}
The online estimation of visual feature prototype is formulated in Equation~\eqref{eq:mu}. For each batch from $\{\boldsymbol{x}_i, y_i\}_{i=1}^B$ from domain $n$, if there are samples from class $c$, then its $\mathbf{v}_c^n$ is updated as below,
\begin{equation}
\begin{aligned}
  \mathbf{v}^n_c|_{new} = 0.5\times \frac{1}{|\Lambda_c|}\sum_{y_i=c}f(\boldsymbol{x}_i) + 0.5\times\mathbf{v}^n_c|_{old},
\end{aligned}
\label{eq:mu}
\end{equation}
where $|\Lambda_c|$ denotes the sample number of class $c$ from current batch.

The covariance $\mathbf{\Sigma}$ is online estimated in the same manner as \cite{wang2019implicit}.

\section{Supplementary Results and Discussions}
\subsection{Further Discussions on Ablation Studies}
Here we added more discussions in terms of our ablation studies. The three core modules in our meta-learning framework were derived from \textcolor{Line}{Equation 1} in the main paper. We perform ablation studies to show the effectiveness of each module design, as well as their complementary benefits to each other. Below we give discussions based on the results in \textcolor{Line}{Tables~2,4} of the main paper.

(1) We proposed distribution-calibrated loss $\mathcal{L}_{dc}$ align the classification loss to a canonical balanced distribution, aiming to handle P(Y) shifts across domains. It outperforms BSCE, as shown in \textcolor{Line}{Table 2-BSCE}.

(2) In addition, $\mathcal{L}_{dc}$ can unify both losses on $\mathcal{D}_{mtr}$ and $\mathcal{D}_{mte}$ to the same balanced distribution, showing better results when applying cross entropy instead in the meta learning setting (\textcolor{Line}{Table 4-d} vs \textcolor{Line}{Table 4-c}). 

(3) Our Visual-Semantic mapping aims to learn domain-aligned unbiased representation by bidirectional Visual-Semantic mapping (\textcolor{Line}{Table 4-e,f,g}) and cross prototype alignment. To validate the effectiveness of performing cross prototype alignment, in \textcolor{Line}{Table 4-k}, we performed ablation study with a uni-domain prototype for alignment, and performance decrease can be noted. Since P(Y) of each domain is different, it is more flexible to build domain-specific prototype for cross-domain alignment, and it simultaneously mitigates the memory bottleneck issue, compared to directly sampling intra-class inter-domain samples for alignment. Moreover, our meta-learning setting can align the feature from $\mathcal{D}_{mte}$ to the prototype of $\mathcal{D}_{mtr}$, which further helps feature alignment across domains.

(4) The feature learned by Visual-Semantic mapping is also important for the augmentation module, otherwise domain shifts may dominate intra-class variances, as demonstrated in \textcolor{Line}{Table 4-h}.

(5) In the augmentation module, we adopted the weighted term {$\boldsymbol{n_k}$} in \textcolor{Line}{Equation 7}. This leads to the situation where ``header" classes would contribute more to updating the covariance matrix of tail classes, whereas ``tailer'' less to head classes. We performed the additional experiments by removing {$\boldsymbol{n_k}$}, with results presented in \textcolor{Line}{Table 4-l}. 

(6) The proposed meta-learning framework integrating the three modules is effective for \ltds{} (\textcolor{Line}{Table 4-i} vs \textcolor{Line}{Table 4-j}).


\subsection{Selection of Different Embeddings}

We utilized typical embeddings as they are available along with previous open-source works on these datasets. Here, we added additional experiments comparing BERT, CLIP, and GloVe on \awalts{}. Results in \tablename~\ref{tab:awa2-embedding} show that the embedding types do not affect the performance much, and our method consistently outperforms Agg w. Embs
\begin{table}[]
\begin{minipage}[b]{0.45\linewidth}
\tabcaption{Results of different embedding types based on Agg and Ours.}
\centering
\label{tab:awa2-embedding}
\setlength{\tabcolsep}{3pt}
\resizebox{\columnwidth}{!}{
\begin{tabular}{lccc}
\toprule
\multicolumn{1}{l}{\textbf{Methods}}  & Acc-U & Acc & H  \\ \midrule
  \multicolumn{1}{l}{Agg} & 26.6 & 31.8 & 37.8 \\ 
  \multicolumn{1}{l}{Agg w. BERT} & 28.2 & 33.1 & 39.1\\
  \multicolumn{1}{l}{Agg w. CLIP} & 27.6 & 32.8 & 39.2 \\
  \multicolumn{1}{l}{Agg w. GloVe} & 27.0 & 32.4 & 38.2\\
 \arrayrulecolor{gray}
 \cmidrule(lr){1-4}
  \multicolumn{1}{l}{Ours w. BERT} & 35.7 & \textbf{42.0} & \textbf{44.6} \\
  \multicolumn{1}{l}{Ours w. CLIP} & \textbf{35.8} & 41.4 & 43.6 \\
  \multicolumn{1}{l}{Ours w. GloVe} & 35.4 & 41.3 & 44.5 \\
\arrayrulecolor{black}
\bottomrule
\vspace{-15pt}
\end{tabular}}
\end{minipage}
\vspace{-10pt}
\hfill
\begin{minipage}[b]{0.45\linewidth}
\tabcaption{Performance changes with different imbalance ratios.}
\centering
\label{tab:awa2-ratio}
\setlength{\tabcolsep}{3pt}
\resizebox{\columnwidth}{!}{
\begin{tabular}{llccc}
\toprule
 \textbf{Ratio} & \multicolumn{1}{l}{\textbf{Methods}}  & Acc-U & Acc & H  \\ \midrule
   \multirow{2}{*}{\textbf{78}} & Agg & 26.6 & 31.8 & 37.8 \\
   & Ours & \textbf{35.7} & \textbf{42.0} & \textbf{44.6}  \\
  \arrayrulecolor{gray}
 \cmidrule(lr){1-5}  \multirow{2}{*}{\textbf{50}} & Agg & 21.7 & 27.6 & 32.8  \\
  & Ours & \textbf{32.0} & \textbf{37.6} & \textbf{41.0} \\ 
 \cmidrule(lr){1-5}    \multirow{2}{*}{\textbf{10}} &
  Agg &  13.9 & 18.2 & 20.7 \\
  & Ours & \textbf{24.0} & \textbf{23.1} & \textbf{32.1} \\  \arrayrulecolor{gray}
\arrayrulecolor{black}
\bottomrule
\vspace{-15pt}
\end{tabular}}
\end{minipage}
\hfill
\end{table}
. 

\subsection{Sample Complexity}
Here, we added experiments with \awalts{} by changing the imbalance ratio of the training set (change head sample number), while kept using the same test set. Results in \tablename~\ref{tab:awa2-ratio} show that ours outperforms Agg counterparts by a large margin in all settings.

\subsection{PACS-ODG}
 
Our targeted problem \ltds{} has a similar setting to open domain generalization (\textbf{ODG}) proposed in \cite{shu2021open}.
We also evaluated our proposed framework on the open domain generalization task on the \pacsodg{} dataset introduced in \cite{shu2021open}. We followed the settings of \cite{shu2021open}. The original class number and domain number of \pacs{}~\cite{li2017deeper} is 7 and 4, respectively. Each domain has its predefined training/validation/testing splits~\cite{li2017deeper}. In the settings of open domain generalization, only part of the label set was selected in each individual training domain, and the trained model is tested on the held-out testing domain consisting of all classes. Following \cite{shu2021open}, three domains are used for training and validation, and the held-out domain is for testing. In line with \cite{shu2021open}, we reported the metrics \textbf{Acc-U} and \textbf{H-U} on the held-out domain. In addition, during testing, we also validated on the testing data of all the domains, where the domain-average accuracy of all non-open classes \textbf{Acc} were reported. The detailed split settings are listed in \tablename~\ref{tab:pacsodg}, and the domain order of each leave-one-domain-out loop is CPS-A, PAC-S, ACS-P, SPA-C. 

Under the same experimental settings, we compared our results of \accu{} and \hu{} with the results of Agg, Epi-FCR~\cite{li2019episodic}, CuMix~\cite{mancini2020towards}, DAML~\cite{mancini2020towards} reported in \cite{shu2021open}. We leveraged the source-code of DAML~\cite{mancini2020towards} and reported its \acc{} result\footnotemark[1]. We also applied state-of-the-art domain generalization algorithm Mixstyle~\cite{zhou2021mixstyle} for comparison. We used the ResNet-18 pretrained from ImageNet as $f$ following previous works~\cite{mancini2020towards,shu2021open}. We did not apply Semantic-Similarity Augmentation module for \pacsodg{} due to the limited class number and the non-existing long-tailed issue of this dataset. 
\footnotetext[1]{similar \textbf{\textit{Acc-U}} and \textbf{\textit{H-U}} results can be achieved as \cite{shu2021open}} 

\begin{table}
\centering
\caption{Settings of Open Domain Generalization of \pacsodg{}.}
\label{tab:pacsodg}
\begin{tabular}{ccc}
   \toprule
   \textbf{Domain}  & \textbf{Training} & \textbf{Testing} \\
   \midrule
   Domain 1  & 0,1,3 & 0,1,2,3,4,5  \\
   Domain 2  & 0,2,4 & 0,1,2,3,4,5  \\
   Domain 3  & 1,2,5 & 0,1,2,3,4,5  \\
   Domain 4  &  -    & 0,1,2,3,4,5,6 \\
   \bottomrule
\end{tabular}
\end{table}

As shown in \tablename~\ref{tab:pacs}, overall our method show favorable performance compared to other methods.  Actually, although the authors of \cite{shu2021open} did not consider the imbalance issue in their work, this indeed exists, since the label set is only partial in each set, leading to the ``infinite'' imbalance ratio. On the other hand, those classes common in all domains would contribute to more samples, thus leading to an overall imbalanced distribution. This imbalance problem, inherent in \textbf{ODG}, was however overlooked in existing studies.

\begin{table}[t]
\centering
\tabcaption{Results on \pacsodg{} dataset.}
\label{tab:pacs}
\setlength{\tabcolsep}{3pt}
\resizebox{\columnwidth}{!}{
\begin{tabular}{lccccccccccccccc}
\toprule
  & \multicolumn{3}{c}{\textbf{Art}} & \multicolumn{3}{c}{\textbf{Sketch}} & \multicolumn{3}{c}{\textbf{Photo}}  & \multicolumn{3}{c}{\textbf{Cartoon}}   & \multicolumn{3}{c}{\textbf{Avg}} \\ \cmidrule(lr){2-4} \cmidrule(lr){5-7} \cmidrule(lr){8-10} \cmidrule(lr){11-13} \cmidrule(lr){14-16}
 \multicolumn{1}{l}{\textbf{Method}} & Acc-U & Acc & H-U & Acc-U & Acc & H-U &      Acc-U & Acc  & H-U & Acc-U & Acc  & H-U & Acc-U & Acc & H-U \\ \midrule
  Agg & 51.4 & - & 38.8 & 49.8 & - & 47.1 & 53.2 & - & 44.2 & 66.4 & - & 49.0 & 55.2 & - & 44.8 \\ 
  Epi-FCR\cite{li2019episodic} &  54.2 & - & 41.2 & 46.4 & - & 46.1 & 70.0 & - & 48.4 & 72.0 & - & 58.2 & 60.6 & - & 48.5 \\
  CuMix\cite{mancini2020towards} & 53.9 & - & 38.7 & 37.7 & - & 28.7 & 65.7 & - & 49.3 & \sec{74.2} & - & 47.5 & 57.9 & - & 41.1 \\ 
  DAML\cite{shu2021open} & 54.1 & \sec{60.6} & 43.0 & \sec{58.5} & \sec{75.5} & \best{56.7} & \sec{75.7} & \sec{68.3} & 53.2 & 73.7 & \sec{76.9} & 54.5 & 65.5 & 70.3 & 51.9 \\
  MixStyle\cite{zhou2021mixstyle} & \sec{56.0} & 57.8 & \sec{47.0} & 51.7 & 75.0 & 44.9 & 58.7 & 57.0 & 33.0 & \best{75.8} & 75.5 & \best{63.6} & 60.5 & 66.3 & 47.1 \\ \midrule
  Ours &  \best{58.4} & \best{66.3} & \best{47.8} & \best{60.4} & \best{80.1} & \sec{49.1} & \best{78.4} & \best{77.2} & \best{71.2}& 71.3 & \best{77.8} & \sec{59.6} & \best{67.1} & \best{75.3} & \best{56.9} \\
\bottomrule
\end{tabular}}
\end{table}


\subsection{Additional Qualitative Results}
In this section, more qualitative results on \awalts{} are shown. 
\subsubsection{Inter-Domain Discrepancy:} 
\begin{figure}[h!]
    \centering
    \includegraphics[width=0.45\linewidth]{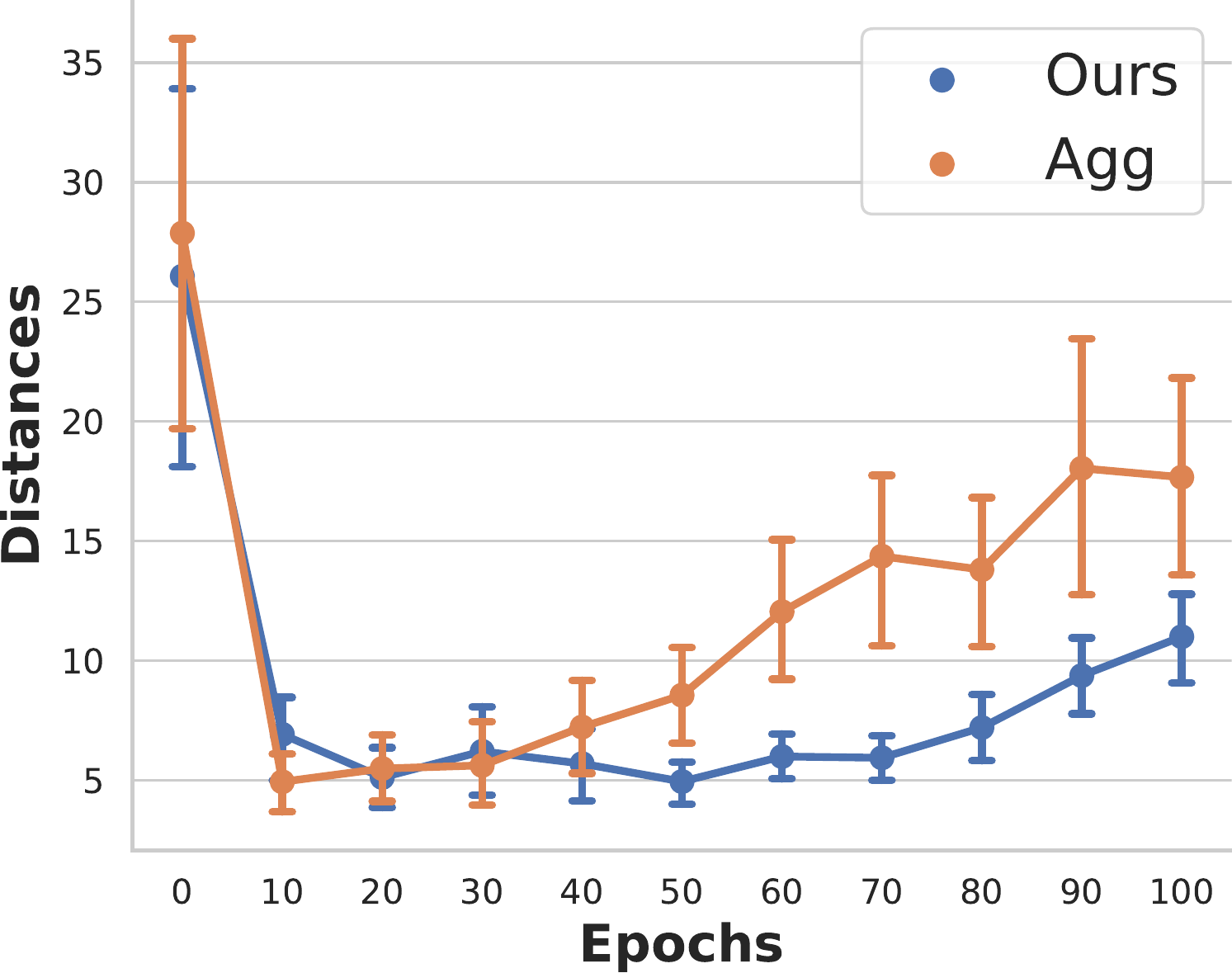}
     \vspace{-15pt}
    \caption{Changes of inter-domain discrepancies on the testing subset (all five domains) during training. The inter-domain discrepancies of Agg start to increase significantly at a very early stage, whereas the discrepancies of our method remain more stable.}
    \label{fig:fd}
\end{figure}

We present the inter-domain discrepancies on the testing subset (all five domains) during the training procedure. The distance is calculated by Fr\'echet distance. It can be observed in \figurename~\ref{fig:fd} that the inter-domain distances were largely decreased and maintained in a small scale by our proposed method, whilst the discrepancies under the Agg baseline are becoming much larger after training. This emphasizes the overfitting on seen domains by the conventional Agg method. 

\subsubsection{Covariance Similarity Before and After Semantic-Similarity Guided Update:}
\begin{figure}[h!]
    \centering
    \includegraphics[width=0.7\linewidth]{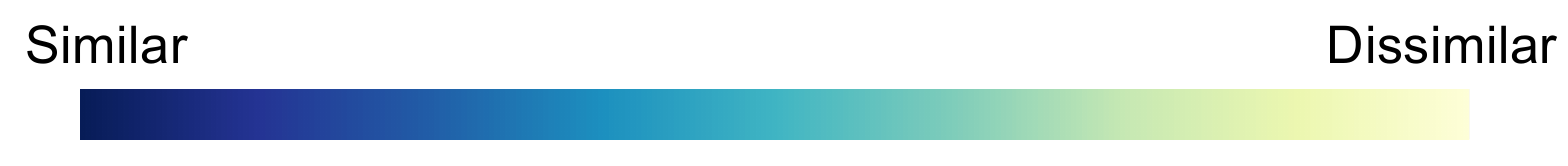}
    \subfigure[Semantic Similarity\label{fig:sim} ]{\includegraphics[width=0.32\linewidth]{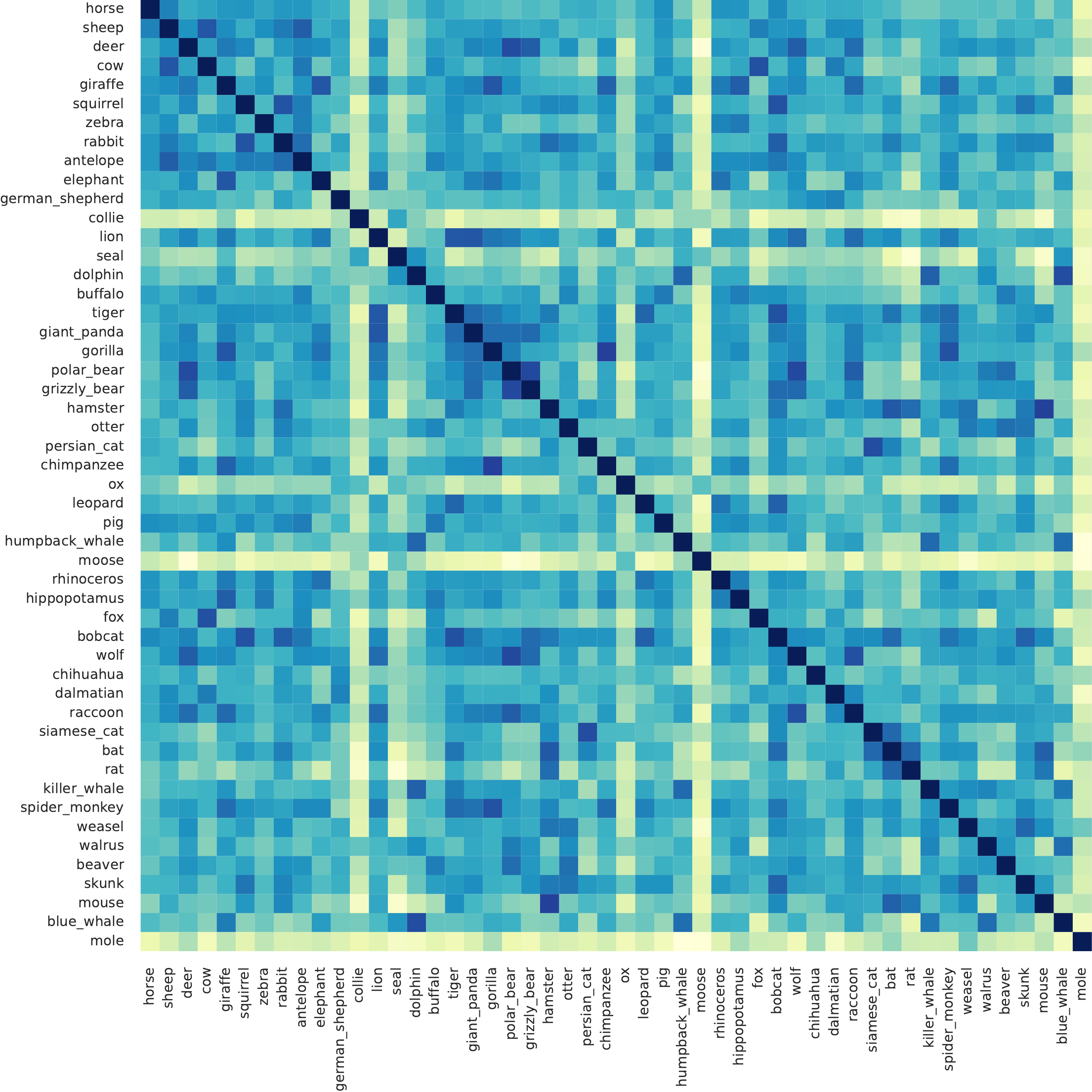}}
    \subfigure[Pairwise Distance of Original Covariances\label{fig:sim_old}]{\includegraphics[width=0.32\linewidth]{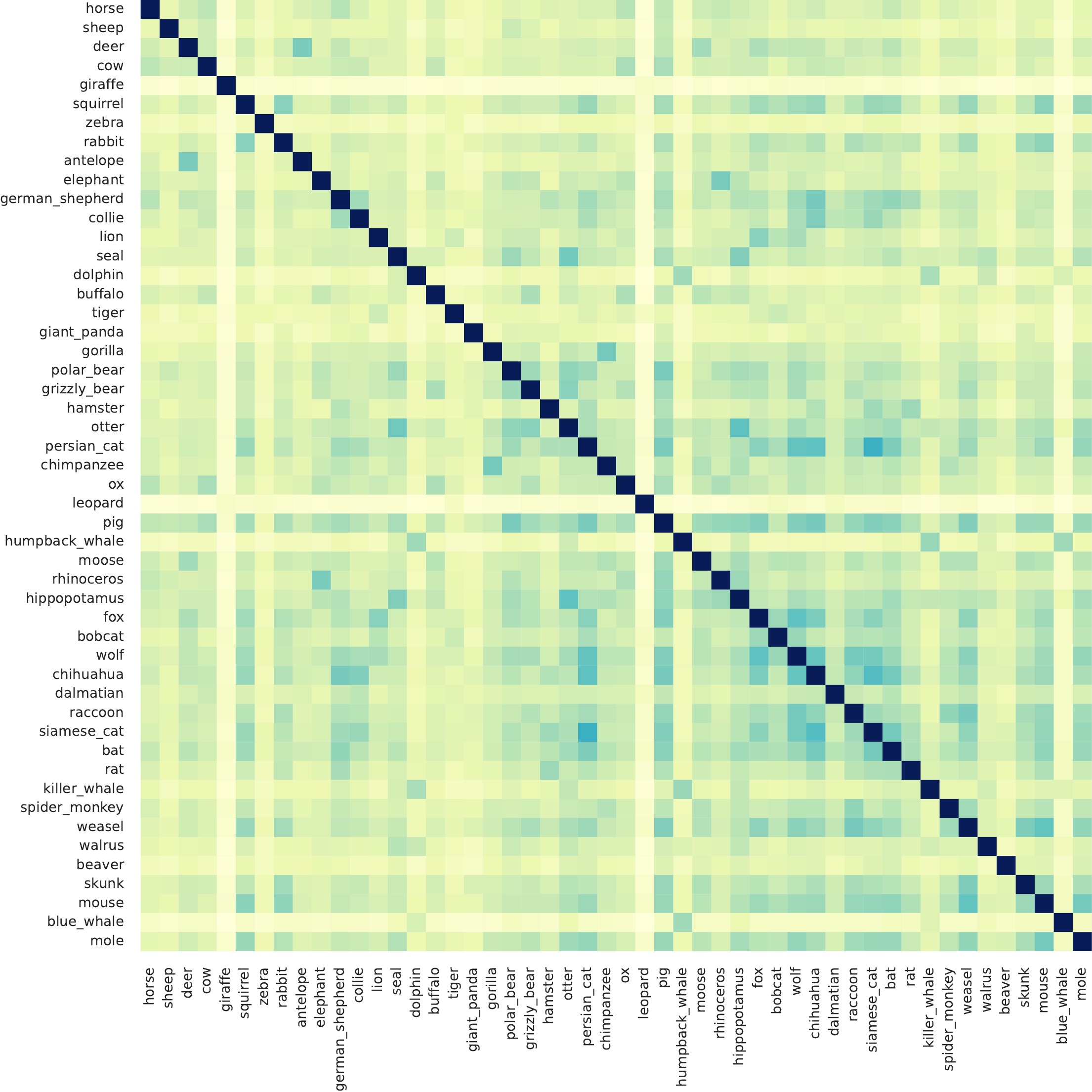}}
    \subfigure[Pairwise Distance of Updated Covariances\label{fig:sim_new}]{\includegraphics[width=0.32\linewidth]{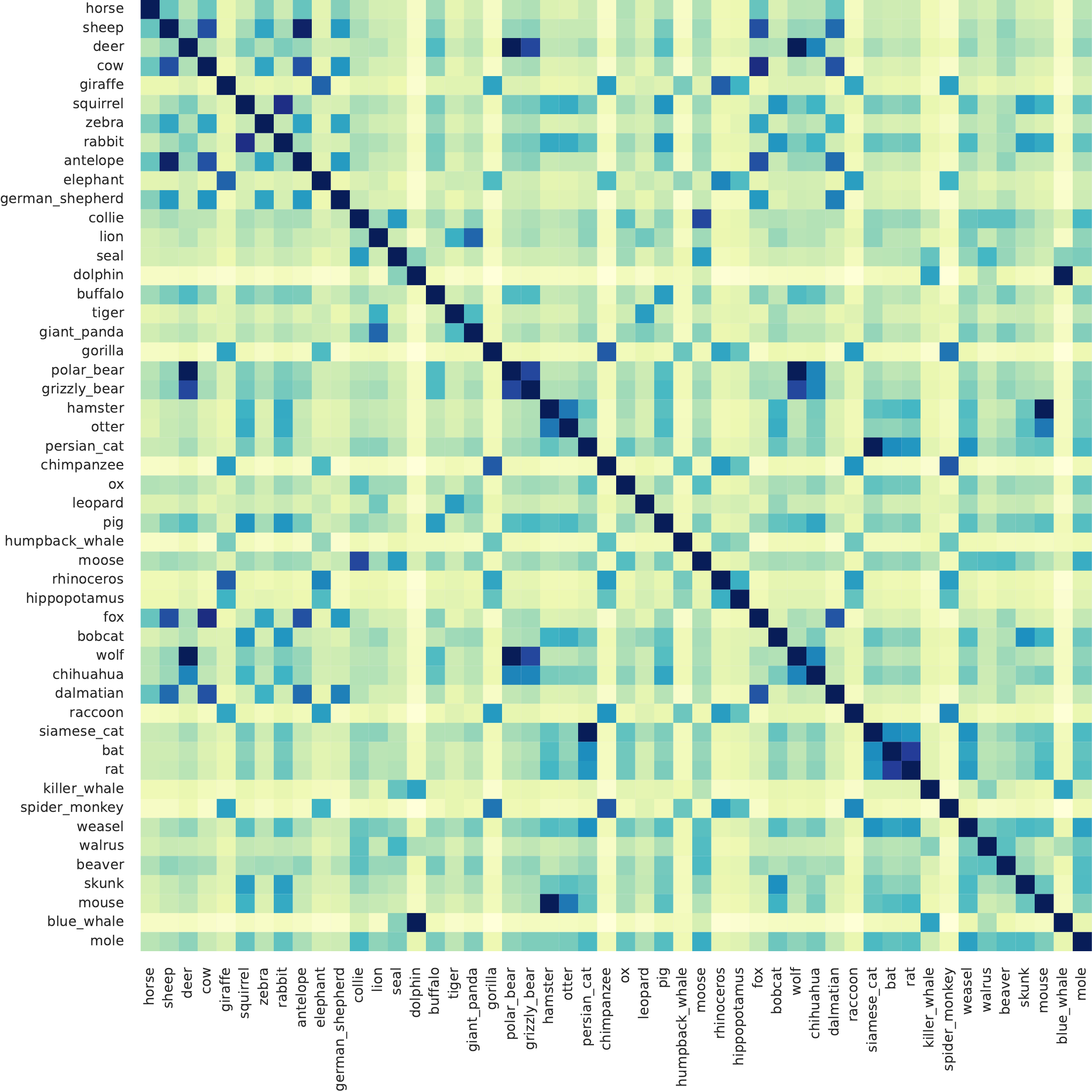}}
    \vspace{-10pt}
    \caption{Pairwise Similarity of Semantic Embeddings, Original and Updated Covariance Matrices. Based on the inter-class semantic  similarity guided from (a), the covariance matrices, especially of those tail classes, are updated based on the statistics from top k most similar classes.}
    \label{fig:sim_matrix}
\end{figure}

\figurename~\ref{fig:sim_matrix} shows the inter-class similarity of the semantic embeddings, as well as of the original and updated covariance matrix. For visualization purposes, \figurename~\ref{fig:sim} shows the cosine similarity of inter-class semantic embeddings, whereas Figs.~\ref{fig:sim_old} and~\ref{fig:sim_new} calculates the pairwise distances between class $i$ and $j$ by $d(i,j)=\exp(-\left \| \mathbf{\Sigma}_i-\mathbf{\Sigma}_j \right \|_2)$. Guided by the similarity derived from semantic embeddings (as in \figurename~\ref{fig:sim}, the original covariance matrix was updated by the top k most similar classes weighted by their class numbers. With the weighted update strategy, the covariance matrices of head classes are not affected too much, whereas those tail covariance matrices can be much influenced by those similar head classes for better modelling. 

\begin{figure}[h!]
    \centering
    \includegraphics[width=\linewidth]{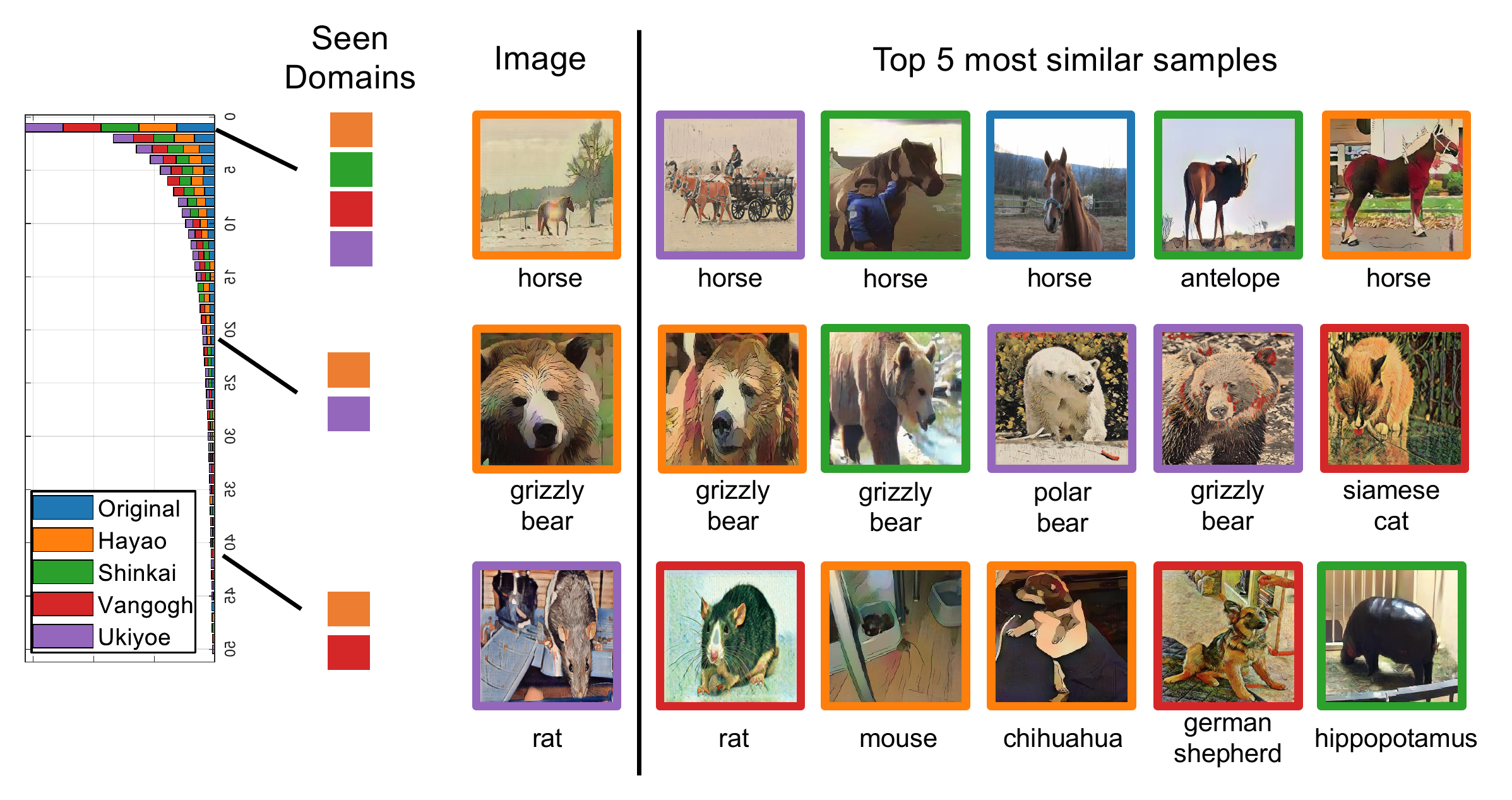}
    \vspace{-15pt}
    \caption{Representative examples of top 5 most similar samples in the semantic embedding space. ``Seen domains'' mean the domains in the training subset of which the current category is available.}
    \label{fig:top5}
\end{figure}

\subsubsection{Visualizations of Top-5 Retrieval}
In \figurename~\ref{fig:top5}, we presented a few examples of top 5 most similar samples in the semantic embedding space when holding Original domain out. We selected one instance from one head, middle, and tail class, separately. The seen domains of each class during training are visualized on the left side. It can be observed from \figurename~\ref{fig:top5} that the similarity is mostly based on its semantic meaning rather than the domain styles. For example, in the Row 1 of \figurename~\ref{fig:top5}, the most similar three samples of the Hayao \textit{horse} are from other different domains. Similarly, in the Row 3, even \textit{rat} of Ukiyoe is not seen during training, the most similar counterpart is from the same class yet a different domain. We also noticed some fine-grained variances between some similar classes based on this visualization, i.e., \textit{grizzly bear} vs \textit{polar bear}, \textit{rat} vs \textit{mouse}. The incapability of distinguishing them may be due to the limited effectiveness of the backbone. One possible direction of future work is to explore more effective and deeper backbones to enable better recognition performance to distinguish such small differences.

\end{document}